\documentclass[twoside]{article}

\usepackage[accepted]{aistats2025}
\usepackage[round]{natbib}

\usepackage{amsmath}
\usepackage{amssymb}
\usepackage{bm}
\usepackage{xcolor}
\usepackage{graphicx}
\usepackage{subfiles}
\usepackage{subcaption}
\usepackage[font=small,labelfont=small]{caption}
\usepackage{hyperref}
\usepackage{yhmath}
\usepackage{multirow}
\usepackage{algorithm}
\usepackage{algorithmic}

\usepackage{import}
\usepackage{xifthen}
\usepackage{pdfpages}
\usepackage{transparent}

\usepackage{booktabs}
\usepackage{subcaption}
\usepackage{pifont}
\newcommand{\cmark}{\ding{51}}%
\newcommand{\xmark}{\ding{55}}%

\usepackage{tikz}
\usetikzlibrary{arrows.meta}

\graphicspath{{../}{./}}

\DeclareMathOperator{\Exp}{Exp}
\DeclareMathOperator{\Log}{Log}
\DeclareMathOperator{\GP}{GP}
\DeclareMathOperator{\WGP}{WGP}
\DeclareMathOperator{\cov}{cov}
\DeclareMathOperator{\vvec}{vec}
\DeclareMathOperator{\diag}{diag}
\DeclareMathOperator{\Tr}{Tr}

\newcommand{\tasks}{\bm{f}}
\newcommand{\dd}{\mathrm{d}}
\newcommand{\feuc}{f_\text{E}}
\newcommand{\pderiv}[2]{\frac{\partial #1}{\partial #2}}
\newcommand{\ppderiv}[3]{\frac{\partial^2 #1}{\partial #2\partial #3}}

\newcommand{\trsp}{\mathsf{T}}  
\newcommand{\ty}[1]{{\scriptscriptstyle{\mathcal{#1}}}}

\newcommand{\riemannsq}{\textsc{Riemann}$^2$ }

\newcommand{\euclideanspace}{\mathbb{R}}
\newcommand{\sphere}{\mathbb{S}}
\newcommand{\SPD}{\mathcal{S}_{\ty{++}}}
\newcommand{\manifold}{\mathcal{M}}

\newcommand{\gaussiandist}[3]{\mathcal{N}(#1;#2,#3)} 

\definecolor{lightred}{rgb}{0.8, 0.22, 0.29}
\definecolor{turquoise}{rgb}{0.25, 0.89, 0.82}
\definecolor{mediumorchid}{rgb}{0.73, 0.33, 0.83}
\definecolor{darkorange}{rgb}{1.0, 0.65, 0.0}
\definecolor{maroon}{rgb}{0.5, 0.0, 0.0}
\definecolor{navy}{rgb}{0.0, 0.0, 0.5}
\definecolor{dodgerblue}{rgb}{0.12, 0.565, 1.0}
\definecolor{crimson}{rgb}{0.86, 0.08, 0.235}
\definecolor{middlegray}{rgb}{0.3, 0.3, 0.3}
\definecolor{lightgray}{rgb}{0.6, 0.6, 0.6}
\DeclareRobustCommand{\dodgerbluecircle}{\tikz{ \filldraw[color=dodgerblue, fill=dodgerblue, thick](0,0) circle (.05);}}
\DeclareRobustCommand{\crimsonline}{\raisebox{2pt}{\tikz{\draw[crimson,solid,line width = 1.1pt](0,0) -- (4mm,0);}}}
\DeclareRobustCommand{\blackline}{\raisebox{2pt}{\tikz{\draw[black,solid,line width = 1.1pt](0,0) -- (4mm,0);}}}
\DeclareRobustCommand{\yellowline}{\raisebox{2pt}{\tikz{\draw[darkorange,solid,line width = 1.1pt](0,0) -- (4mm,0);}}}
\DeclareRobustCommand{\maroonline}{\raisebox{2pt}{\tikz{\draw[maroon,solid,line width = 1.1pt](0,0) -- (4mm,0);}}}
\DeclareRobustCommand{\grayellipse}{\tikz{ \filldraw[color=middlegray, fill=lightgray!60, thick](0,0) ellipse (.15 and 0.08);}}

\begin{document}

%

%
\runningauthor{Leonel Rozo, Miguel González-Duque, Noémie Jaquier, Søren Hauberg}

\twocolumn[

\aistatstitle{Riemann$^2$: Learning Riemannian Submanifolds from Riemannian Data}

\aistatsauthor{Leonel Rozo$^*$ \And Miguel González-Duque$^*$ \And Noémie Jaquier \And Søren Hauberg}

\aistatsaddress{ Bosch center for AI \And  DBtune \And  KTH  \And Technical Uni. of Denmark} ]

\begin{abstract}
Latent variable models are powerful tools for learning low-dimensional manifolds from high-dimensional data. 
However, when dealing with constrained data such as unit-norm vectors or symmetric positive-definite matrices, existing approaches ignore the underlying geometric constraints or fail to provide meaningful metrics in the latent space.
To address these limitations, we propose to learn Riemannian latent representations of such geometric data.
To do so, we estimate the pullback metric induced by a Wrapped Gaussian Process Latent Variable Model, which explicitly accounts for the data geometry.
This enables us to define geometry-aware notions of distance and shortest paths in the latent space, while ensuring that our model only assigns probability mass to the data manifold.
This generalizes previous work and allows us to handle complex tasks in various domains, including robot motion synthesis and analysis of brain connectomes.
\vspace{-0.2cm}
\end{abstract}

\vspace{-0.2cm}
\section{Introduction}
\label{sec:intro}
\vspace{-0.2cm}
Geometrically-constrained data appears in many application domains such as biology~\citep{Macaulay23:HyperbolicPhylogenetic}, robotics~\citep{Urain22:LieAlgebraDS,Rozo22:OrientationProMPs}, motion modeling~\citep{Jaquier2024:GPHLVM,He24:RiemannianSDF}, and medical imaging~\citep{Pennec20:RiemannMedicalImaging,Baust20:ManifoldDataMedicalImage}. For example, the orientation of rigid bodies is commonly represented by unit quaternions in the hypersphere $\sphere^3$, or diffusion tensor images are encoded in the space of symmetric positive-definite matrices $\SPD^d$. One of the main challenges when analyzing or predicting this type of data arises when finding a low-dimensional subspace in a high-dimensional setting, as classical techniques are not tailored to non-Euclidean data. However, recent works on Gaussian processes latent variable models (GPLVM)~\citep{Mallasto19:RiemannianGPLVM} and variational autoencoders (VAE)~\citep{miolane:rvae:2020,hadi:rss:2021}, offer compelling solutions to this problem by generating data complying with the given geometric constraints.\looseness=-1 

A key aspect the aforementioned works still overlook is the data manifold structure.  
In other words, the assumption that the latent space is Euclidean often fails at capturing the underlying data manifold~\citep{Shao:RiemannianGeometryOfGenerativeModels:2018,Hauberg19:OnlyBS}. This problem was recently tackled by endowing the latent space of generative models with a Riemannian metric induced by the non-linear mapping of the model decoder~\citep{Tosi:UAI:2014,arvanitidis:iclr:2018,Park23:DiffModelPullback}. Importantly, this metric encapsulates the data uncertainty into the latent space, and therefore enables downstream tasks that comply with the data manifold geometry.
This proved useful in reinforcement learning~\citep{Tennenholtz22:PullbackMetricRL}, protein sequencing~\citep{detlefsens:proteins:2022}, and robotics~\citep{Chen18:ActiveLearningMetrics,Scannell21:LatentGeodesicRobot,hadi:rss:2021}, among other disciplines. \looseness=-1

Inspired by the foregoing works on the geometry of generative models and the need of handling data with geometric constraints in downstream tasks, we here tackle the problem of learning low-dimensional representations of Riemannian data that pull back the data space metric into the latent space. 
In other words, \textbf{this paper} proposes a generative model that learns Riemannian submanifolds from Riemannian data, hereinafter referred to as \riemannsq\!\!. 
Formally, our model learns a stochastic latent variable model $f\colon\mathcal{Z}\to\manifold$, which maps latent embeddings to Riemannian data lying on $\manifold$. The latent space $\mathcal{Z}$ is equipped with the pullback metric induced by the decoder $f$, similarly to~\citet{Tosi:UAI:2014}. This way, we immerse the latent space, via the nonlinear mapping $f$, into an ambient Riemannian manifold. Specifically, we employ the geometry-aware wrapped GPLVM (WGPLVM)~\citep{Mallasto19:RiemannianGPLVM} to represent $f$, so that the decoded latent embeddings are guaranteed to lie on $\manifold$. Unlike~\citet{Tosi:UAI:2014} and~\citet{Mallasto19:RiemannianGPLVM}, \riemannsq builds on multitask Gaussian Processes (GPs)~\citep{Bonilla:multitaskGP:2007}, which are crucial to account for correlated outputs. This is relevant to account for correlations in products of manifolds, often found in, e.g., synchronized motion generation.\looseness=-1

In summary, \textbf{our contributions are}: (1) We define latent space geometries for multitask WGPLVMs; (2) we derive an expression for the distribution of the Riemannian pullback metric for multitask wrapped GPs, thus generalizing the work of~\citet{Tosi:UAI:2014}; (3) we define back-constraints for WGPLVMs by leveraging Riemannian kernels, accounting for the Riemannian geometry of the observation space; and (4) we formulate the wrapped GP likelihood to account for the change of volume of the distribution. Our experiments confirm the importance of considering both the data intrinsic geometry and its distribution.\looseness=-1

\begin{table}[]
    \centering
    \resizebox{\columnwidth}{!}{%
    \begin{tabular}{lcccc}
        \toprule
         & UQ & Riemannian Data & Pullback metric & Output correlation \\
        \midrule
        \cite{arvanitidis:iclr:2018} & \xmark & \xmark & \cmark & \xmark \\
        \cite{hadi:rss:2021} & \xmark & (\cmark) & \cmark & \xmark \\
        \cite{Tosi:UAI:2014} & \cmark & \xmark & \cmark & \xmark \\
        \cite{Mallasto19:RiemannianGPLVM} & \cmark & \cmark & \xmark & \xmark \\
        \cite{miolane:rvae:2020} & \cmark & \cmark & \xmark & \xmark \\
        \midrule
        \riemannsq & \cmark & \cmark & \cmark & \cmark \\ 
        \bottomrule
    \end{tabular}%
    }
    \caption{Comparison of \riemannsq against previous works. 
    Note that the manifold-aware components of~\citep{hadi:rss:2021} are specific to the hypersphere $\sphere^3$.} 
    \label{tab:intro:comparison-between-methods}
    \vspace{-0.4cm}
\end{table}
\vspace{-0.2cm}
\section{Related Work}
\vspace{-0.3cm}
Latent generative models like GPLVMs~\citep{Lawrence03:GPLVM,Titsias10:BayesGPLVM} and VAEs~\citep{Kingma:VAEs:2014} generally assume a Gaussian prior distribution over the mapping function or the latent variables. Interestingly, as shown by~\citet{Tosi:UAI:2014,Shao:RiemannianGeometryOfGenerativeModels:2018,arvanitidis:iclr:2018}, the decoding function of these models induces a pullback Riemannian metric on their latent space, opening the door to a better understanding on the learned embeddings geometry. However, none of these works considered data lying on Riemannian manifolds. Several generative models such as GMMs~\citep{jaquier:manipulability:2021}, normalizing flows~\citep{Rezende20:NFsToriSphere,Chen24:RiemannianFM}, or diffusion models~\citep{Huang22:RiemannianDifussion} were formulated to account for the intrinsic geometry of the data. However, latent generative models with geometry-aware decoders are still scarce, and few recent works stand out: the Wrapped GPLVM~\citep{Mallasto19:RiemannianGPLVM} and the geometry-aware VAEs~\citep{miolane:rvae:2020,hadi:rss:2021}. We take inspiration from them in this work.\looseness=-1

Table~\ref{tab:intro:comparison-between-methods} summarizes how our approach contributes to the state of the art. \citet{Tosi:UAI:2014} and~\citet{arvanitidis:iclr:2018} define latent space geometries using GPLVMs and VAEs, respectively. However, their decoders are not manifold-aware, and thus assign probability mass outside the specified manifold in the ambient space. \cite{Mallasto19:RiemannianGPLVM} defined GPLVMs on tangent bundles, thus restricting the mass to a given manifold, but did not pull the data geometry back onto the latent space. Similarly, \cite{miolane:rvae:2020} design a VAE using a Riemannian normal distribution for the decodings, but do not equip the latent space with a metric. \citet{hadi:rss:2021} define latent space geometries of a hypersphere VAE, but rely on handcrafted uncertainty estimates. $\textsc{Riemann}^2$ is manifold-aware, achieves automatic uncertainty quantification (UQ) through the use of GPLVMs, and defines latent space geometries. Furthermore, it accounts for correlated outputs via multitask kernels.\looseness=-1
\vspace{-0.2cm}

\begin{figure*}
    \centering
    \def\svgwidth{1.6\columnwidth}
    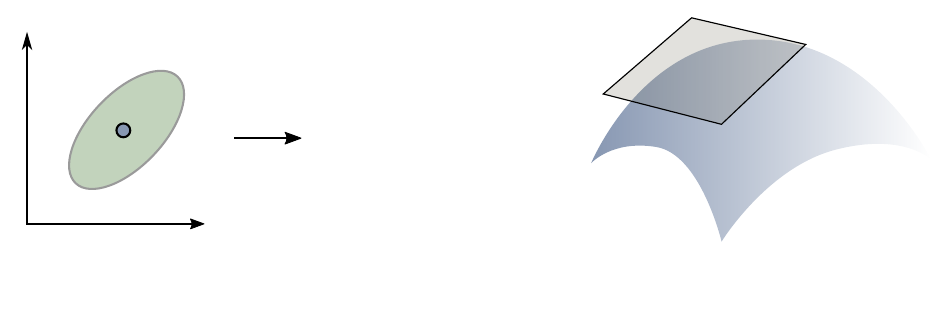

    \caption{\riemannsq: To learn a Riemannian submanifold from Riemannian data, our method pulls back a Riemannian metric $\tilde{g}_{\bm{x}}$ to a latent space via a Wrapped GPLVM. In this model, each latent code $\bm{x}\in\euclideanspace^L$ defines a distribution of tangent vectors $\feuc(\bm{x})\sim \GP(\bm{0}, k)$, which is then pushed forward onto the manifold $\manifold$ via the exponential map $\Exp_{b(\cdot)}$. Our framework enables geodesics that, when decoded, comply with the data manifold and are guaranteed to lie on $\manifold$.}
    \vspace{-0.3cm}
    \label{fig:my_label}
\end{figure*}

\section{Background}
\label{sec:background}
\vspace{-0.3cm}
\paragraph{Gaussian Process Latent Variable Model:}
A GPLVM defines a generative mapping from latent variables $\bm{x}_i\in\euclideanspace^Q$ to data $\bm{y}_i\in\euclideanspace^D$ by modeling the corresponding non-linear transformation $f$ with a multitask Gaussian process (GP)~\citep{Rasmussen, Lawrence03:GPLVM}.
The data is assumed to be normally distributed, i.e., $y_{i,d} \sim \gaussiandist{y_{i,d}}{f_{i,d}}{\sigma^2_d}$ with,  
\begin{equation}
 f_{i,d} \sim \GP(m_d (\bm{x}_i), k^{\bm{x}}_{d}(\bm{x}_i,\bm{x}_i)) \;\;\text{and}\;\; \bm{x}_i \sim \mathcal{N}(\bm{0},\bm{I}),
\label{eq:GPLVM}
\end{equation}
where $y_{i,d}$ is the $d$-th dimension of $\bm{y}_i$, $m_d:\euclideanspace^Q\to \euclideanspace$, and $k^{\bm{x}}_d\!:\!\euclideanspace^Q \!\times\!\euclideanspace^Q\!\to\!\euclideanspace$ are the GP mean and kernel function, and $\sigma_d^2$ is a hyperparameter.
The main design choice in a GP is the kernel $k^{\bm{x}}_d$, which encodes the covariance between two predictions $f(\bm{x}_i), f(\bm{x}_j)$ as a similarity measure between the corresponding inputs $\bm{x}_i, \bm{x}_j$.
A common choice is the square exponential (SE) kernel parametrized by a variance $\sigma^2$ and a lengthscale $\theta$. \looseness=-1

A common approach to model vector-valued functions $f: \euclideanspace^Q \to \euclideanspace^D$ is to learn each dimension $f_d$ independently with its own GP~\citep[Sec. 3.3]{Alvarez:kernels_for_multioutput:2012}.
However, this disregards possible correlations between the output dimensions.
We consider a more general formulation proposed by~\cite{Bonilla:multitaskGP:2007}, in which such correlations are modeled using a learned positive-definite matrix $k^{\tasks}$. The covariance thus becomes,
\begin{equation}
    \label{eq:background:multitask-gp-kernel}
    \cov(f_{i, r}, f_{j, s}) = k\big(f_r(\bm{x}_i), f_s(\bm{x}_j)\big) = k^{\tasks}_{rs}k^{\bm{x}}(\bm{x}_i, \bm{x}_j).
\end{equation}
In other words, $k = k^{\tasks}\otimes k^{\bm{x}}$, where $\otimes$ is the Kronecker product. 
The GPLVM hyperparameters and latent variables are often optimized using \textit{maximum likelihood} or \textit{maximum a posteriori} (MAP) estimates. 
The task covariance can be parametrized by a low-rank square root $\bm{B}$ added to a diagonal variance vector $k^{\tasks} = (\bm{B}\bm{B}^\top + \diag(\bm{v}))$ \citep{Daskalakis:low_rank_approximations_of_gps:2022,Gardner:GPyTorch:2018}.
In large-datasets settings, contemporary methods use inducing points and variational approximations of the evidence~\citep{Titsias10:BayesGPLVM}. 
Note that, unlike neural-networks, GPLVMs are data efficient and provide automatic uncertainty quantification, a relevant aspect when estimating a stochastic Riemannian metric, as explained in Sec.~\ref{sec:method:pullback_metric}.\looseness=-1

\paragraph{Riemannian Geometry:}
\label{sec:background:riemannian}
A smooth manifold $\manifold$ is a topological space that is locally Euclidean, meaning that a neighborhood surrounding a point $\bm{p}\in\manifold$ is diffeomorphic to $\euclideanspace^M$. 
We can locally approximate $\manifold$ at each point $\bm{p} \in \manifold$ with a tangent space $\mathcal{T}_{\bm{p}} \manifold \simeq \euclideanspace^M$, defined as the set of derivatives of all smooth curves that pass through $\bm{p}$~\citep{Lee00Smooth}. 
The collection of all tangent spaces is called the \textit{tangent bundle}, and is formally defined as the disjoint union $\mathcal{T}\manifold = \sqcup_{\bm{p}\in \manifold} \mathcal{T}_{\bm{p}}\manifold$. 

Given a function $h\colon \manifold\to \mathcal{N}$ between two smooth manifolds, the \textit{differential} at $\bm{p}\in \manifold$ is the linear function $\dd h_{\bm{p}}\colon \mathcal{T}_{\bm{p}}\manifold \to \mathcal{T}_{h(\bm{p})}\mathcal{N}$ that maps tangent vectors $\bm{v}_{\bm{p}}\in \mathcal{T}_{\bm{p}}\manifold$ to tangent vectors $\dd h_{\bm{p}}(\bm{v}_{\bm{p}})\in\mathcal{N}$. 
When considering coordinates $\varphi=(x^{1}, \dots, x^{M})$ around $\bm{p}$ and $\psi=(y^{1},\dots,y^{N})$ around $h(\bm{p})$, the differential $\dd h_{\bm{p}}$ is represented as the Jacobian matrix,\footnote{We omit the choice of charts from our notation of $[\dd h_{\bm{p}}]$, but we make this choice explicit when necessary.}
\begin{equation}
    \label{eq:background:jacobian_of_h}
[\dd h_{\bm{p}}] = \bm{J}_h(\varphi(\bm{p})) = \left[\left.\pderiv{(y^{i}\circ h)}{x^{j}}\right|_{\varphi(\bm{p})}\right]_{i, j = 1}^{N, M} .
\end{equation}
Intuitively, the Jacobian~\eqref{eq:background:jacobian_of_h} transforms tangent vectors on $\manifold$ to tangent vectors on $\mathcal{N}$, which allows us to define latent space geometries as discussed later.

A Riemannian manifold $(\manifold, g)$ is a smooth manifold equipped with a \textit{Riemannian metric} $g_{\bm{p}}$, i.e., a smoothly-varying inner product over $\mathcal{T}_{\bm{p}}\manifold$~\citep{Lee18Riemann}. 
Given a smooth curve $\gamma: [a,b]\to \manifold$, its Riemannian length is given by $\text{L}[\gamma] = \int_a^b \sqrt{g_{\gamma(t)}(\dot{\gamma}(t), \dot{\gamma}(t))} \mathrm{d}t$.
\textit{Geodesics} are defined as curves that locally minimize this length.
Our method deals with Riemannian submanifolds, which are locally-Euclidean topological subspaces $\mathcal{S}\subseteq\manifold$ that inherit the metric of $\manifold$ via their immersion. 
Specifically, the immersion $f \colon \mathcal{S}\to (\manifold, g)$ induces a \textit{pullback} metric 
$\tilde{g}_{\bm{x}}$ on $\mathcal{S}$ which, for $\bm{x}\in \mathcal{S}$ and $\bm{v}_1, \bm{v}_2 \in \mathcal{T}_{\bm{x}}\mathcal{S}$, is given by~\citep[Chap. 2]{Lee18Riemann},
\begin{equation}
    \label{eq:background:pullback_definition}
    \tilde{g}_{\bm{x}}(\bm{v}_1, \bm{v}_2) = g_{f(\bm{x})}\big(\dd f_{\bm{x}}(\bm{v}_1), \dd f_{\bm{x}}(\bm{v}_2)\big).
\end{equation}
Intuitively the pullback metric $\tilde{g}_{\bm{x}}$ evaluates on tangent vectors of $\mathcal{T}_{\bm{x}}\mathcal{S}$ by ``moving'' them to $\mathcal{T}_{f(\bm{x})}\manifold$ to compute their inner product. 
\riemannsq defines pullback metrics in the latent spaces of manifold-aware latent variable models by explicitly computing an approximation of $\dd f_{\bm{x}}$, where $f$ is a GPLVM defined on manifolds. 

To operate with data on Riemannian manifolds, we additionally leverage the Euclidean tangent spaces. 
To do so, we resort to mappings back and forth between $\mathcal{T}_{\bm{p}}\manifold$ and $\manifold$. 
The exponential map $\Exp_{\bm{p}}(\bm{u})\colon \mathcal{T}_{\bm{p}}\manifold \to \manifold$ maps a point $\bm{u}\!\in\!\mathcal{T}_{\bm{p}}\manifold$ to a point $\bm{y}\!\in\!\manifold$, so that it lies on the geodesic $\gamma$ satisfying $\gamma(0)\!=\!\bm{p}$ and $\gamma'(0)\!=\!\bm{u}\!\in\!\mathcal{T}_{\bm{p}}\manifold$ at time $1$.
The exponential map $\Exp_{\bm{p}}(\bm{u})$ establishes a \emph{local} diffeomorphism around $\bm{p}$, whose inverse function is the logarithmic map $\Log_{\bm{p}}(\bm{y}): \manifold \to \mathcal{T}_{\bm{x}}\manifold$. 
Intuitively, the logarithm map $\Log_{\bm{p}}(\bm{y})$ represents the direction and speed at which we need to ``shoot'' a geodesic based at $\bm{p}$ to land at $\bm{y}$.\looseness=-1

\paragraph{Wrapped GPLVMs:}
Introduced by~\citet{Mallasto19:RiemannianGPLVM}, Wrapped GPLVMs (WGPLVM) extend GPLVMs to data lying on Riemannian manifolds.
WGPLVMs build on a \textit{Wrapped GP} (WGP)~\citep{Mallasto18:WGPsRiemannian}) defined on the tangent spaces of $\manifold$.
To fit a WGP on a dataset $\{(\bm{x}_i, \bm{y}_i)\}\in\euclideanspace^Q\times \manifold$, we first project the observations onto the tangent bundle $\mathcal{T}\manifold$, constructing a new tangent space dataset $\{(\bm{x}_i, \Log_{b(\bm{x}_i)}(\bm{y}_i))\}$, where $b\!:\!\euclideanspace^Q\!\to\!\manifold$ assigns a \textit{basepoint} on the manifold $\manifold$ to each latent variable $\bm{x}$. 
By identifying $\mathcal{T}_{b(\bm{x})}\manifold$ with $\euclideanspace^M$, this dataset is now Euclidean, and we can fit a GP $\feuc \sim \GP(\bm{0}, k)$ to it. 

To predict over unseen points $\bm{x}_*$, we first compute the Euclidean posterior, 
and then push it forward onto $\manifold$ via $\Exp_{b(\bm{x}_*)}$.
In summary, a WGP is $\Exp_{b(\cdot)}\big(\feuc(\cdot)\big)$, where $\feuc\sim\GP(\bm{0}, k)$ is a Euclidean GP learned on the tangent spaces $\mathcal{T}_{b(\cdot)}\manifold$. 
Analogously, WGPLVMs are defined as WGPs in which the latent variables $\bm{x}_i$ are unobserved and thus optimized w.r.t the likelihood of the data $\bm{y}_i$ alongside the GP hyperparameters.\looseness=-1

\section{\riemannsq}
\label{sec:riemannsq}
\vspace{-0.2cm}
We here introduce \riemannsq to learn Riemannian submanifolds via WGPLVMs, that is, to learn latent representations of Riemannian data alongside a Riemannian pullback metric in the latent space. 
From a bird's eye view, we first learn a mapping $f:\euclideanspace^Q\to (\manifold, g)$, which we then consider as a stochastic immersion, allowing us to pull back the metric $g$ onto the latent space $\euclideanspace^Q$ using Eq.~\eqref{eq:background:pullback_definition}. 
We define the mapping $f$ as a WGPLVM. Instead of a WGP that considers each output dimension independently, like~\citet{Mallasto18:WGPsRiemannian}, we propose a multitask WGP that explicitly models the covariance across output dimensions. 
The Jacobian of this WGP is then used to approximate a distribution of the differential $\dd f_{\bm{x}}$ and compute the metric $\tilde{g}_{\bm{x}}$.

Formally, let $f = \Exp_{b(\cdot)}\circ\, \feuc: \euclideanspace^Q\to (\manifold, g)$ be a multitask WGP. The differential $\dd f_{\bm{x}}$ is computed by applying the chain rule as,
\begin{equation}
    \dd f_{\bm{x}} = \dd (\Exp_{b(\cdot)}\circ\, \feuc)_{\bm{x}}
    = (\dd \Exp_{b(\bm{x})})_{\feuc(\bm{x})}(\dd \feuc)_{\bm{x}} . \label{eq:method:differential_of_wgp}
\end{equation}
Therefore, we need to compute the differential of the exponential map w.r.t.\@ the tangent GP $\feuc$, and the derivative of $\feuc$ w.r.t.\@ the latent variable $\bm{x}$. In coordinates, these correspond to Jacobians, as discussed in Sec.~\ref{sec:background:riemannian}.
Next, we compute the distribution of the Euclidean Jacobian $\bm{J}_{\feuc}(\bm{x}) = [(\dd \feuc)_{\bm{x}}]$ under the correspondence $\mathcal{T}_{b(\bm{x})}\manifold\simeq\euclideanspace^M$. 
In Sec.~\ref{sec:method:pullback_metric} we then obtain a point estimate of the pullback metric $\tilde{g}$ by approximating $(\dd \Exp_{b(\bm{x})})_{\feuc(\bm{x})}$ using the posterior mean of $\feuc$. 
Section~\ref{sec:method:vectorization_and_practicalities} discusses the problem of operating with tangent spaces and bundles in practice. 
Finally, training and additional priors on the latent space of \riemannsq are introduced in Sec.~\ref{sec:method:training} and~\ref{sec:method:priors}, respectively.
\riemannsq is summarized in Algorithm~\ref{alg:riemann2-training}.\looseness=-1

\vspace{-0.2cm}
\subsection{Jacobian of a Multitask Euclidean GP}
\label{sec:method:jacobian_of_multitask_gp}
\vspace{-0.2cm}
Following~\citet[Chap. 9.4]{Rasmussen}, the derivative of a GP is another GP as long as its covariance function is differentiable. This also holds for multitask Euclidean GPs $\feuc$ with covariance $k = k^{\tasks}\otimes k^{\bm{x}}$ (see App.~\ref{app:PullbackDistribution}). 
Given the dataset $\{\bm{x}_i, \feuc(\bm{x}_i)\}_{i=1}^N\subseteq \euclideanspace^Q\times \euclideanspace^M$, this implies that the joint distribution of the data and the transpose Jacobian $\bm{J}_{\feuc(\bm{x}_*)}^\top$ is of the form,
\begin{equation}
\label{eq:method:joint_distribution_of_data_and_jacobian}
    \begin{bmatrix}
    \vvec(\bm{F}) \\[0.1cm]
    \vvec\left(\bm{J}_{\feuc}^\top(\bm{x}_*)\right)
    \end{bmatrix} \sim \mathcal{N}\left(\bm{0}, \begin{bmatrix}
    \bm{K} & \partial \bm{K} \\
    \partial \bm{K}^\top & \partial^2 \bm{K} 
    \end{bmatrix}\right) ,
\end{equation}
where we defined $\bm{F}\in\euclideanspace^{N\times M}$ in vector form as $\vvec(\bm{F})\!=\![f_1(\bm{x}_1),\dots,f_1(\bm{x}_N),\dots,f_m(\bm{x}_1),\dots,f_m(\bm{x}_N)]^\top$, $\bm{K}$ is the $NM\times NM$ matrix representing the covariance between the elements in $\vvec(\bm{F})$, and $\partial \bm{K}$, $\partial^2 \bm{K}$ are the first and second derivatives of $\bm{K}$ with respect to the latent variable $\bm{x}$, which we derive next for $k = k^{\tasks}\otimes k^{\bm{x}}$.

Intuitively, differentiating a kernel defined as a Kronecker product resembles the product rule. 
Namely, for $\bm{K} = k^{\tasks}\otimes \bm{K}^{\bm{x}}$ with $\bm{K}^{\bm{x}} = \left[k^{\bm{x}}(\bm{x}_n, \bm{x}_{a})\right]_{n, a = 1}^{N}$, we have, 
\begin{align*}
    \partial \bm{K} &= k^{\tasks}\otimes \partial \bm{K}^{\bm{x}}, \partial \bm{K}^{\bm{x}} = \left[\pderiv{k^{\bm{x}}(\bm{x}_n, \bm{x}_*)}{x^{(r)}}\right]_{n, r = 1}^{N,Q} ,\\
    \partial^2 \bm{K} &= k^{\tasks}\otimes \partial^2 \bm{K}^{\bm{x}}, \partial^2 \bm{K}^{\bm{x}} = \left[\ppderiv{k^{\bm{x}}(\bm{x}_*, \bm{x}_*)}{x^{(r)}}{x^{(s)}}\right]_{r, s = 1}^{Q} .
\end{align*}
Notice that the task kernel $k^{\tasks}$ is factored out as it does not depend on $\bm{x}$ (see App.~\ref{sec:appendix:matrix_computations_of_multitask_kernel_derivatives} for details).

\begin{algorithm}[tb]
   \caption{\riemannsq}
   \label{alg:riemann2-training}
\begin{algorithmic}
\small
   \STATE {\bfseries Input:} Observations $\{\bm{y}_n\}_{n=1}^N$  with $\bm{y}_n\in\manifold$ , prior on hyperparameters $p(\bm{\psi})$.
   \STATE {\bfseries Output:} Latent variables $\{\bm{x}_n\}_{n=1}^N$, hyperparameters $\bm{\psi}$ including the kernel hyperparameters, point estimates $\mathbb{E}[\tilde{\bm{G}}]$ of the Riemannian pullback metric.
   \STATE {\bfseries Initialization:}
   \STATE \hspace{\algorithmicindent} Set the prior distribution $p(\bm{x})$.
   \STATE \hspace{\algorithmicindent} Initialize the latent variables $\{\bm{x}_n\}_{n=1}^N$.
   \STATE {\bfseries Training via MAP:}
   \begin{ALC@g}
   \REPEAT
   \STATE Compute the WGP marginal likelihood~\eqref{eq:WGP-marginal-likelihood}.
   \STATE $\bm{X},\psi \gets \mathsf{OptStep}(\log p(\bm{y}|\bm{x}))$.
   \UNTIL{convergence}
   \end{ALC@g}
   \STATE {\bfseries Pullback metric:}
   \STATE \hspace{\algorithmicindent} Compute the distribution~\eqref{eq:method:posterior_matrix_normal_of_jacobian} of $\bm{J}_{\feuc}$.
   \STATE \hspace{\algorithmicindent} Compute point estimates $\mathbb{E}[\tilde{\bm{G}}]$ using~\eqref{eq:method:point_estimate_of_pullback}.
\end{algorithmic}
\end{algorithm}

Using the joint distribution~\eqref{eq:method:joint_distribution_of_data_and_jacobian}, the posterior over the Jacobian is computed as a usual GP posterior as,
\begin{align}
    \label{eq:method:vec_jacobian_distribution}
    \vvec\left(\bm{J}_{\feuc}^\top(\bm{x}_*)\right) \sim&\,\mathcal{N}\!\left(\vvec(\partial \bm{K}^{\bm{x}\top}(\bm{K}^{\bm{x}})^{-1}\bm{F}),\right. \\ \notag
    &\left.k^{\tasks} \otimes (\partial^2\bm{K}^{\bm{x}} - \partial \bm{K}^{\bm{x}\top} (\bm{K}^{\bm{x}})^{-1}\partial \bm{K}^{\bm{x}})\right).
\end{align}
The posterior~\eqref{eq:method:vec_jacobian_distribution} is equivalent to the following matrix normal distribution~\citep[Chap. 2.]{Gupta_Nagar:matrix_valued_dist:1999},
\begin{subequations}
\label{eq:method:posterior_matrix_normal_of_jacobian}
\begin{align}
    \bm{J}_{\feuc}^\top(\bm{x}_*) \sim &\,\mathcal{MN}_{Q\times M}(\partial \bm{K}^{\bm{x}\top}(\bm{K}^{\bm{x}})^{-1}\bm{F}, \label{eq:method:jacobian_mean_posterior} \\
    &\partial^2\bm{K}^{\bm{x}} - \partial \bm{K}^{\bm{x}\top} (\bm{K}^{\bm{x}})^{-1}\partial \bm{K}^{\bm{x}}, \label{eq:method:jacobian_cov_over_rows} \\
    &k^{\tasks}). \label{eq:method:jacobian_cov_over_columns}
    \end{align}
\end{subequations}
where Eqs.~\eqref{eq:method:jacobian_mean_posterior},~\eqref{eq:method:jacobian_cov_over_rows}, and~\eqref{eq:method:jacobian_cov_over_columns} are the posterior mean, covariance over rows, and covariance over columns, respectively. A more detailed derivation is provided in App.~\ref{sec:appendix:distribution_of_multitask_gp}.
We argue that this result makes intuitive sense: The usual Jacobian posterior for the dimension-independent case appears as the posterior mean and covariance over rows (i.e., over the data), while the posterior covariance over columns is precisely the task covariance.
In summary, the posterior distribution of the Jacobian of a multitask GP is the matrix normal distribution in Eq.~(\ref{eq:method:posterior_matrix_normal_of_jacobian}). 
Next, we leverage this distribution to compute an estimate of the pullback metric.\looseness=-1 

\subsection{Pullback Metric of a Wrapped GP}
\label{sec:method:pullback_metric}
\vspace{-0.2cm}
Given the distribution~\eqref{eq:method:posterior_matrix_normal_of_jacobian} of the differential of the Euclidean part of a WGP, the missing component to compute $\dd f_{\bm{x}}$ in Eq.~\eqref{eq:method:differential_of_wgp} is the differential of the exponential map.
Since computing the posterior distribution of $(\dd \Exp_{b(\bm{x})})_{\feuc(\bm{x})}$ is involved and manifold-specific, we propose a deterministic approximation given by the mean posterior of the Euclidean GP $\feuc$. 
After evaluating the exponential map on the mean posterior $\widehat{\feuc}(\bm{x})$, the differential can be computed using autodifferentiation.
We denote this approximation as $\bm{J}_{\Exp_{b(\bm{x})}}\big(\widehat{\feuc}(\bm{x})\big)$. 
We also approximate the metric evaluations $g_{f(\bm{x})}$ in Eq.~\eqref{eq:background:pullback_definition} by using the posterior mean of the Wrapped GP $f$, whose matrix representation we denote by $\widehat{\bm{G}}(\bm{x})$.

By applying the foregoing approximations, we derive the pullback metric $\tilde{g}_{\bm{x}}$ of Eq.~\eqref{eq:background:pullback_definition} in matrix form as,
\begin{equation}
    \label{eq:method:matrix_expression_of_pullback}
    \tilde{\bm{G}} = \bm{J}_{\feuc}^\top \bm{J}_{\Exp_{b(\bm{x})}}^\top \widehat{\bm{G}} \bm{J}_{\Exp_{b(\bm{x})}}\bm{J}_{\feuc} = \bm{J}_{\feuc}^\top \check{\bm{G}} \bm{J}_{\feuc}.
\end{equation}
Finally, we leverage the distribution of the posterior Jacobian of the Euclidean GP~\eqref{eq:method:posterior_matrix_normal_of_jacobian} to compute a point estimate of the Riemannian pullback metric $\tilde{\bm{G}}$. By using properties of the matrix normal distribution~\citep[Theorem 2.3.5 (ii)]{Gupta_Nagar:matrix_valued_dist:1999}, we obtain
\begin{align}
\begin{split}
    \label{eq:method:point_estimate_of_pullback}
    \mathbb{E}[\tilde{\bm{G}}] &= \mathbb{E}[\bm{J}_{\feuc}^\top]\,\check{\bm{G}}\,\mathbb{E}[\bm{J}_{\feuc}] + \Tr(\check{\bm{G}}^\top \bm{K}^{\tasks})\,\bm{\Sigma}_{\text{r}}(\bm{J}_{\feuc}^\top) ,
\end{split}
\end{align}
where $\bm{\Sigma}_{\text{r}}$ is the posterior covariance over rows given by Eq.~\eqref{eq:method:jacobian_cov_over_rows} (see also App.~\ref{app:PullbackDistribution}). 
Importantly, the point estimate of $\tilde{\bm{G}}$ in Eq.~\eqref{eq:method:point_estimate_of_pullback} unlocks the computation of geodesics in the latent space. 
It is worth noting that Eq.~\eqref{eq:method:point_estimate_of_pullback} generalizes the work of \citet{Tosi:UAI:2014}, whose approach exclusively considers Euclidean data, i.e., $\check{\bm{G}}=\bm{I}$, and dimension-independent GPs with no correlation between output dimensions, i.e., $k^{\tasks}=\bm{I}$.
\looseness=-1

\subsection{How to Represent Tangent Spaces}
\label{sec:method:vectorization_and_practicalities}
\vspace{-0.2cm}
When considering WGPs, the specific representation of the tangent space $\mathcal{T}_{\bm{p}}\manifold$ is often left aside, although it is highly relevant when using WGPs in practice.
This relates to the need of constructing smooth frames, i.e., collections of smoothly-varying bases over the tangent bundle $\mathcal{T}_b\manifold=\sqcup_{\bm{x}\in\euclideanspace^Q}\mathcal{T}_{b(\bm{x})}\manifold$, where $b$ is the basepoint function~\citep{Mallasto18:WGPsRiemannian,Mallasto19:RiemannianGPLVM}. 
We here discuss how we implement the concepts of tangent space and bundle, and how it differs from alternative implementations.

Current software~\citep{Townsend:pymanopt:2016,miolane:geomstats:2020} implements logarithmic maps $\Log_{\bm{p}}(\bm{u})$ by considering their output vectors as elements of the ambient space, instead of relying on an intrinsic formulation w.r.t.\@ the coordinates of a basis. 
For example, with the sphere $\sphere^2\subseteq \euclideanspace^3$, tangent vectors at $\bm{p}\in\sphere^2$ are considered as $3$-dimensional vectors belonging to the plane orthogonal to $\bm{p}\in\euclideanspace^3$. 
There exist two ways to think about $\mathcal{T}_{\bm{p}}\sphere^2$ when implementing a WGP on $\sphere^2$: \emph{(1)} as an abstract vector space, where we place a GP prior on the coefficients of a chosen basis; \emph{(2)} as a subspace of $\euclideanspace^3$ whose vectors we regress directly in ambient space~\citep{miolane:geomstats:2020}.
We find the latter interpretation potentially problematic, as in practice, current implementations learn a projected version of WGPs~\citep{Myers:geomstatschallenge:2022}, where the learned tangent vectors do not necessarily belong to the tangent space. 
This problem compounds when considering manifolds such as the space of symmetric positive-definite matrices $\SPD^k$, where current implementations treat tangent vectors as $k\times k$ symmetric matrices instead of abstract vectors.

We here consider the former alternative: We define a vectorization for each $\mathcal{T}_{\bm{p}}\manifold$. 
Specifically, instead of considering $\Log_{\bm{p}}(\bm{q})$ as a vector in the ambient space where $\manifold$ is embedded in, we specify a basis of $\mathcal{T}_{\bm{p}}\manifold$ and define the output of $\Log_{\bm{p}}(\bm{q})$ as a linear combination of this basis. 
For example, in the case $\manifold=\sphere^2$, this implies learning a tangent GP on a $2$-dimensional space instead of a $3$-dimensional one.
In the case of $\SPD^k$, this implies regressing $k(k-1)/2$ output dimensions instead of $k^2$. 
For most manifolds, such a basis (a.k.a. \textit{frame}) cannot be built smoothly and globally.
Nevertheless, we found that non-smooth choices (see App.~\ref{app:tangentspaces}) work well in practice.\looseness=-1 

\subsection{A Note on Training Wrapped Models}
\label{sec:method:training}
\vspace{-0.2cm}
As mentioned in Sec.~\ref{sec:background}, the GPLVM latent variables and hyperparameters are often inferred by maximizing the marginal log-likelihood $\log p(\bm{y} | \bm{x})$ or the MAP estimate $\log \big( p(\bm{y} | \bm{x}) p(\bm{x}) \big)$.
In the case of wrapped models, \citet{Mallasto18:WGPsRiemannian, Mallasto19:RiemannianGPLVM} instead maximized the marginal likelihood of the tangent space vectors $\bm{v}_i=\Log_{b(\bm{x}_i)}(\bm{y}_i)$ alongside the GP parameters.
However, as the exponential map pushes forward the Euclidean distribution onto $\manifold$, the marginal likelihood of a WGP must consider the change of volume induced by $\Exp_{b(\cdot)}$. Dismissing it may lead to ill-optimized densities. We instead account for this change in volume using the change of variable formula (see App.~\ref{app:wgp}) and optimize the WGP marginal log-likelihood,
\begin{equation}
\vspace{-0.3cm}
    \label{eq:WGP-marginal-likelihood}
    \log p(\bm{y}|\bm{x}) = \log \mathcal{N}\big(\bm{v}; \bm{0}, \bm{K}\big) - \log \det \left(\frac{\partial \Exp_{b(\bm{x})}}{\partial \bm{v}}\right).
    \vspace{-0.2cm}
\end{equation}

\begin{figure*}
    \centering
    \begin{subfigure}[tbp]{.9\textwidth}
        \centering
        \includegraphics[trim={0.0cm 1.5cm 0.0cm 1.5cm},clip,width=0.23\textwidth]{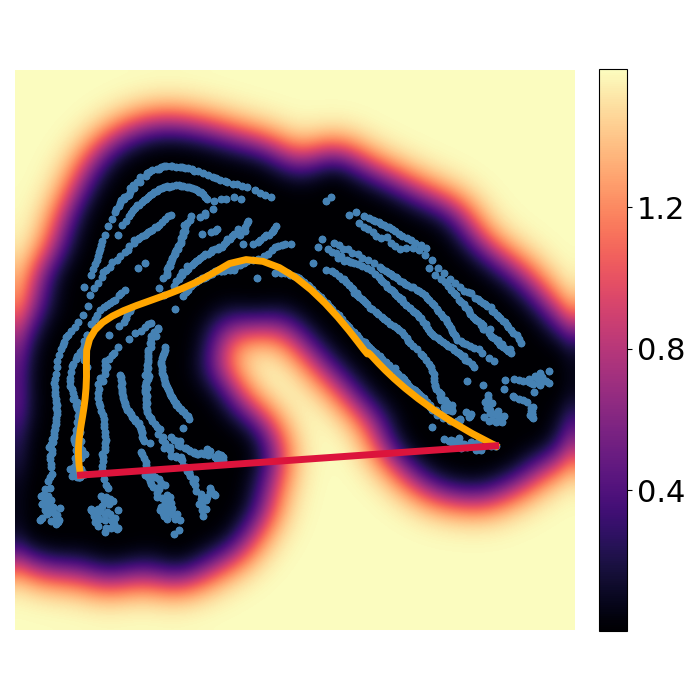}
        \includegraphics[trim={0.0cm 1.5cm 0.0cm 1.5cm},clip,width=0.23\textwidth]{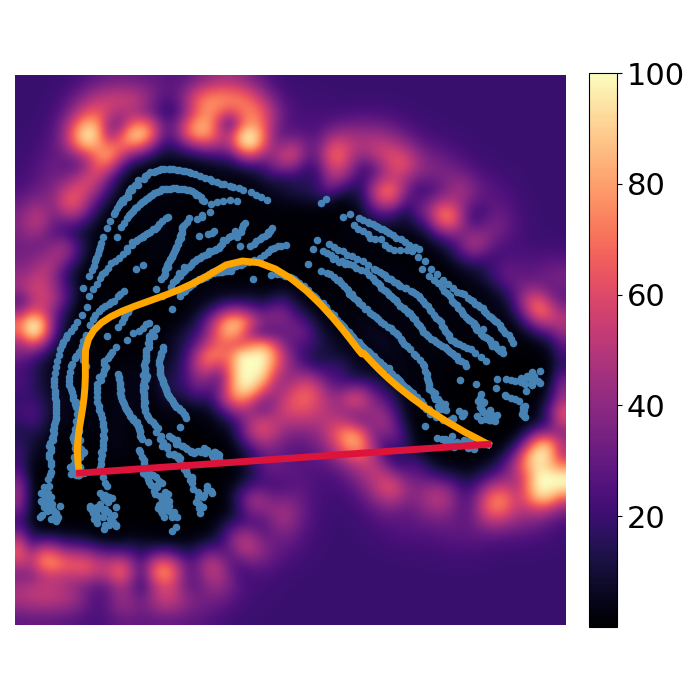}
        \includegraphics[trim={0.0cm 1.0cm 0.0cm 1.0cm},clip,width=0.23\textwidth]{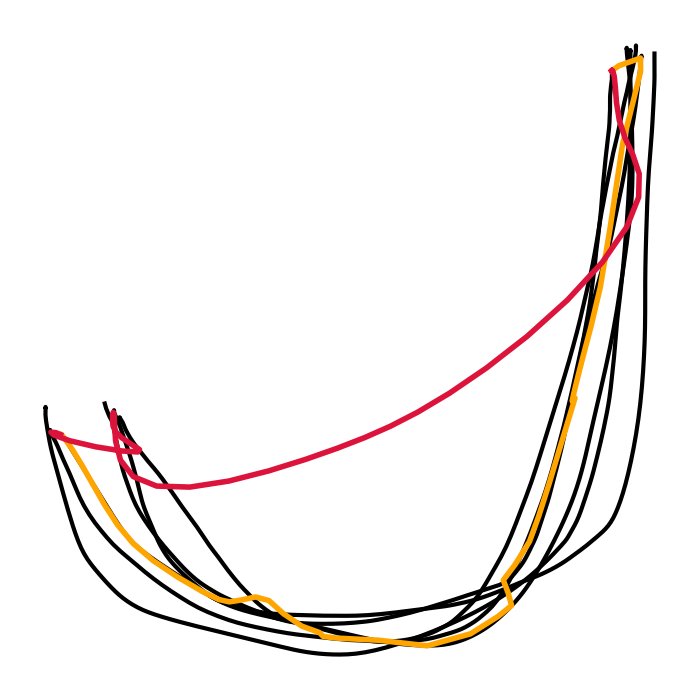}
        \includegraphics[trim={5.0cm 0.0cm 5.0cm 0.0cm},clip, width=0.23\textwidth]{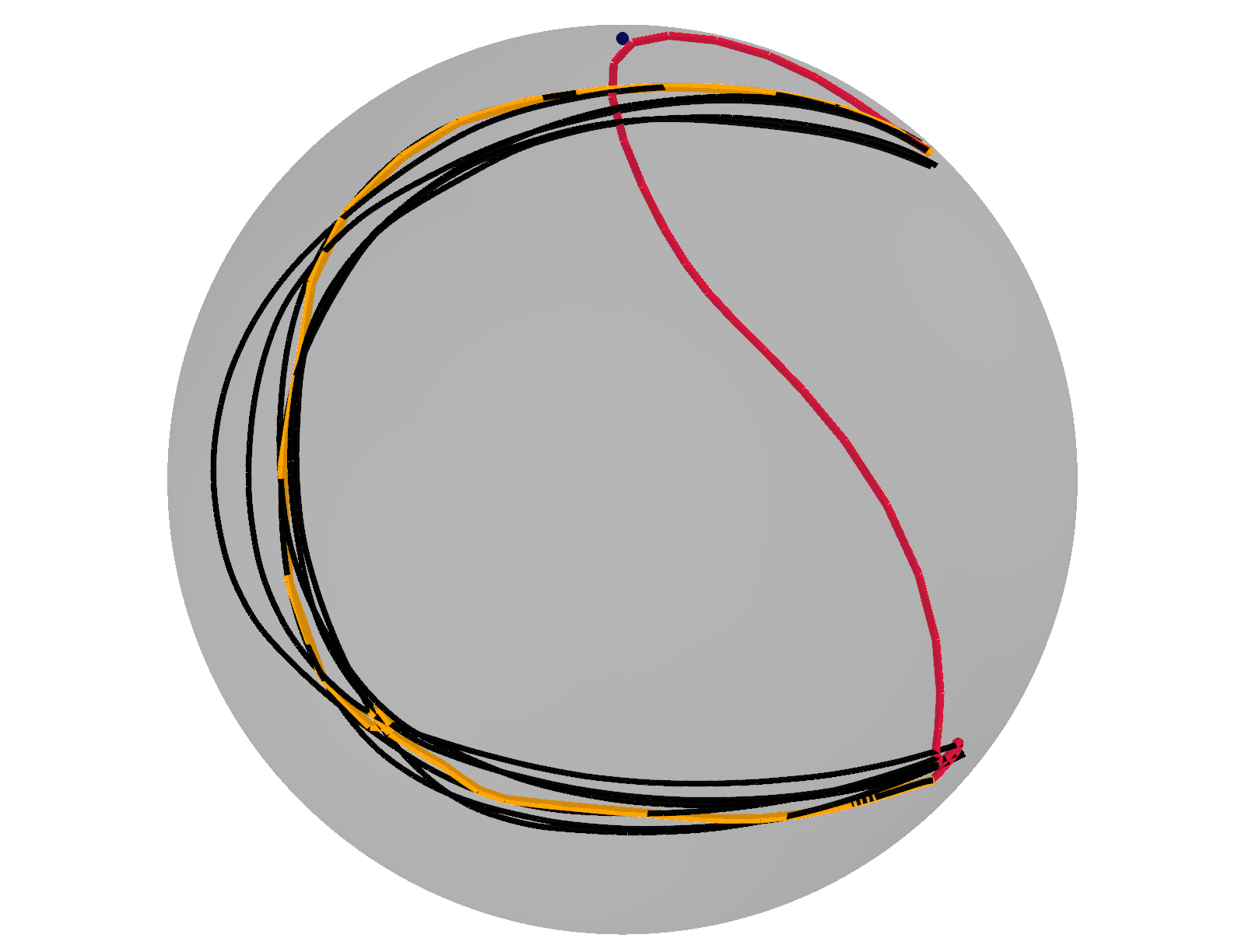}
        \caption{WPGLVM and \riemannsq latent space with one (decoded) geodesic each (\crimsonline, \yellowline).}
    \end{subfigure}%
    
    \begin{subfigure}[b]{.9\textwidth}
        \centering
        \includegraphics[trim={0.0cm 1.5cm 0.0cm 1.5cm},clip,width=0.23\textwidth]{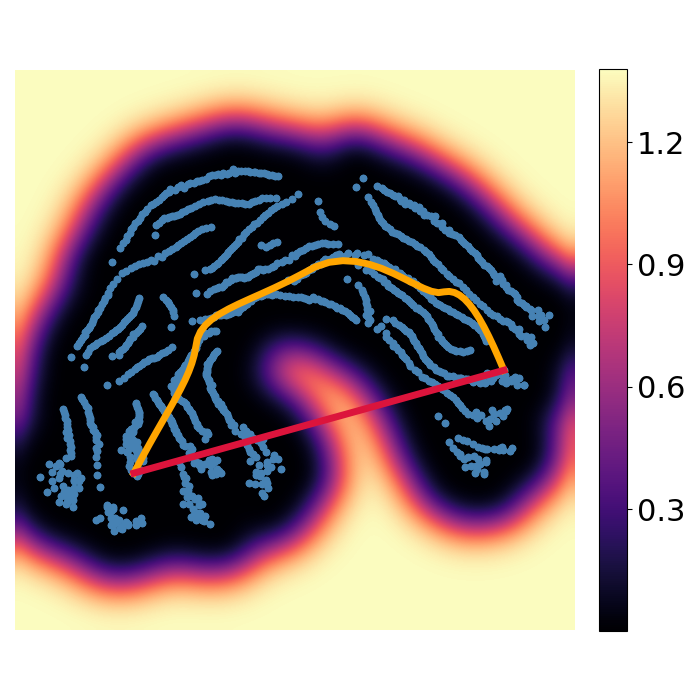}
        \includegraphics[trim={0.0cm 1.5cm 0.0cm 1.5cm},clip,width=0.23\textwidth]{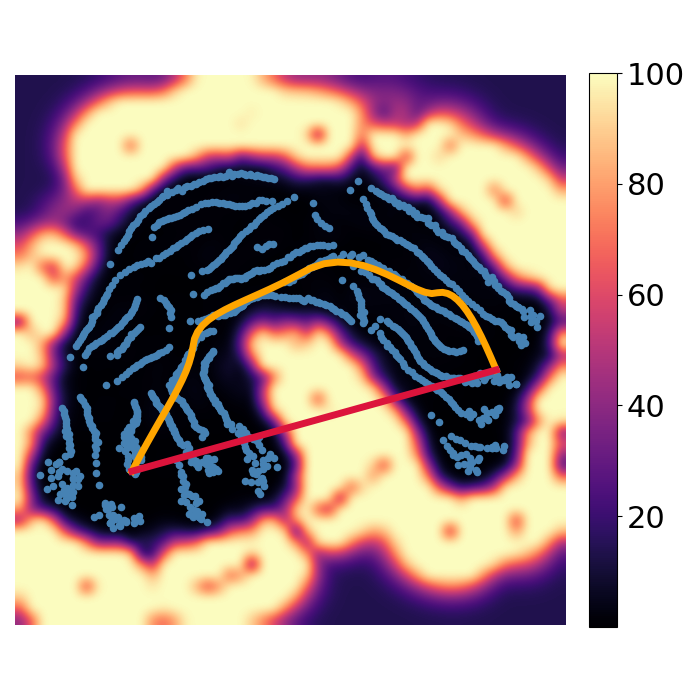}
        \includegraphics[trim={0.0cm 1.0cm 0.0cm 1.0cm},clip,width=0.23\textwidth]{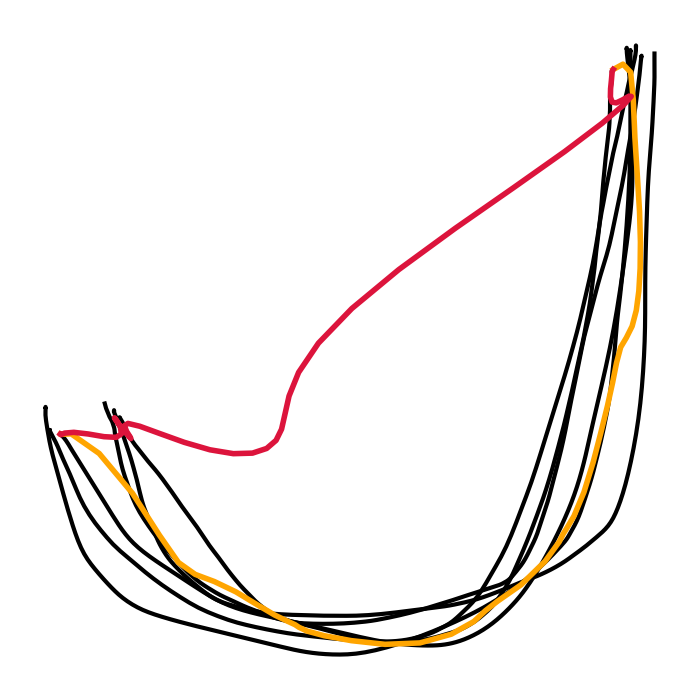}
        \includegraphics[trim={5.0cm 0.0cm 5.0cm 0.0cm},clip,width=0.23\textwidth]{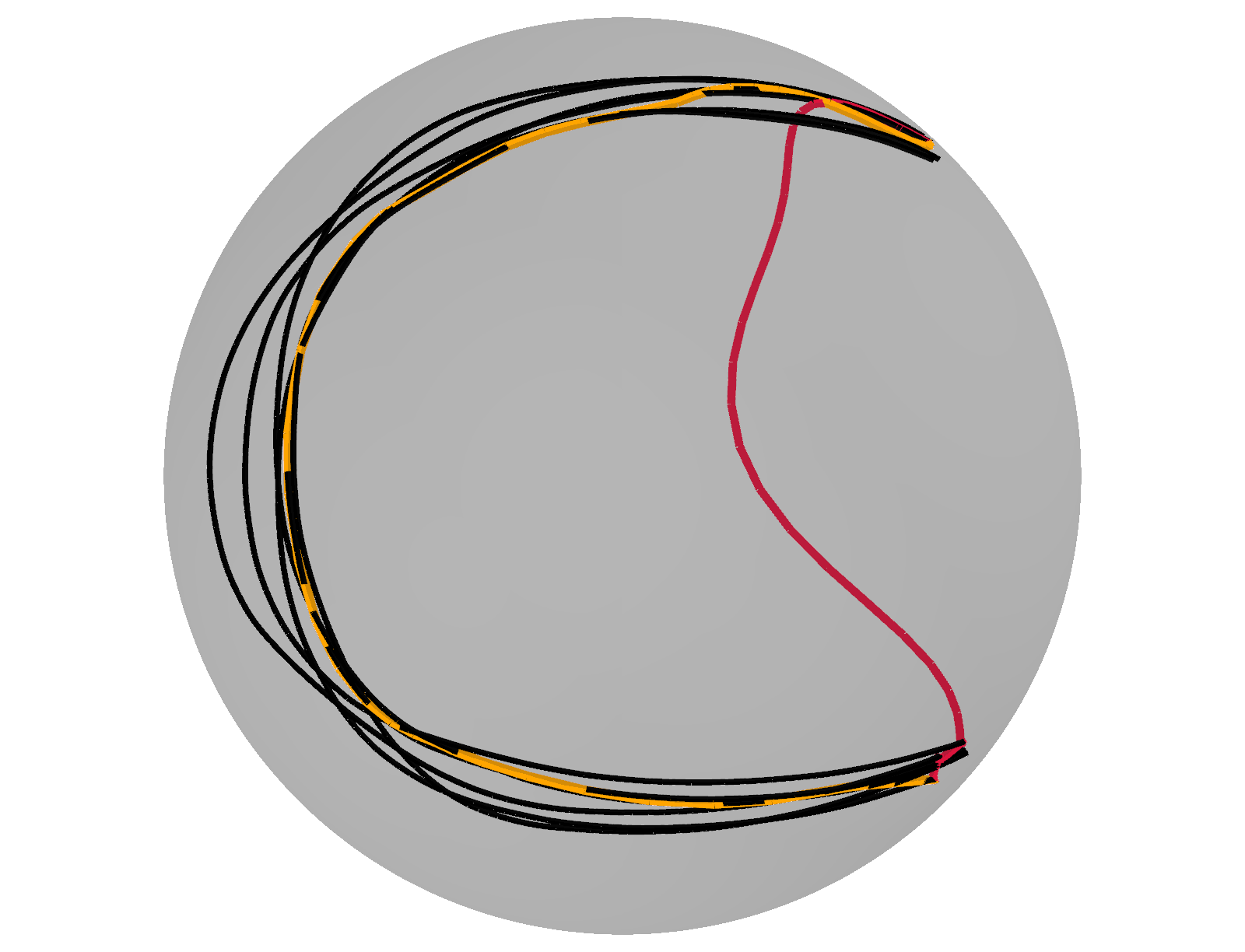}
        \caption{GPLVM and pGPLVM latent space with one (decoded) geodesic each (\crimsonline, \yellowline).}
    \end{subfigure}%
    \caption{Illustrative example on $\euclideanspace^2 \times \sphere^2$: From \emph{left} to \emph{right}: Latent variables (\dodgerbluecircle) with model uncertainty and magnification factor of the pullback metrics, demonstrations (\blackline) and decoded geodesics (\crimsonline, \yellowline) on $\euclideanspace^2$ and $\sphere^2$. 
    }
    \vspace{-0.4cm}
    \label{fig:experiments:toy_experiment}
\end{figure*}

\subsection{Priors and Back Constraints}
\label{sec:method:priors}
\vspace{-0.2cm}
Priors on the latent space and back constraints~\citep{Lawrence06:BackConstrGPLVM,Urtasun08:TopologicalGPLVM} are often used to bias the structure of the latent space. Similarly, the latent space of \riemannsq can be augmented with priors $p(\bm{x})$. 
For instance, in Sec.~\ref{sec:experiments}, we use the Gaussian process dynamical model (GPDM) prior~\citep{Wang08:GPDM} to account for the temporal structure of the data.
\riemannsq can also be augmented with back constraints that define the latent variables as function of the data via the mapping,
\begin{equation}
    x_{i,q} = \sum_{j=1}^N w_{q,j} k^{\manifold}(\bm{y}_i,\bm{y}_j).
\label{eq:backconstraints}
\end{equation}
Importantly, the mapping~\eqref{eq:backconstraints} must be manifold-aware, which we achieve by expressing data similarities via a Riemannian kernel $k^{\manifold}$, following the formulations in~\citep{Borovitskiy20:GPManifolds,Azangulov23:NonCompactKernels}.\looseness=-1


\vspace{-0.2cm}
\section{Experiments}
\label{sec:experiments}
\vspace{-0.2cm}
We test \riemannsq on the following experiments: \emph{(1)} an illustrative example with synthetic data on $\euclideanspace^2\times\sphere^2$, \emph{(2)} a robot end-effector motion synthesis experiment on $\euclideanspace^3\times\sphere^3$, \emph{(3)} a manipulability learning scenario in $\euclideanspace^2\times\SPD^2$, and \emph{(4)} brain connectomes in $\SPD^{15}$.
We compare \riemannsq against three GP-based baselines: \emph{(1)} a vanilla GPLVM, which is manifold-unaware and does not equip the latent space with a metric; \emph{(2)} the pullback GPLVM (pGPLVM) from~\citet{Tosi:UAI:2014}, which assumes Euclidean data but endows the latent space with a pullback metric through a dimension-independent GP; and \emph{(3)} a multitask variant of the WGPLVM proposed in~\citep{Mallasto19:RiemannianGPLVM}, which is manifold-aware but does not pullback the data manifold geometry in the latent space. 
All models are trained and tested under the same settings. The latent spaces of WPGLVM and \riemannsq differ from the latent spaces of GPLVM and pGPLVM due to the intrinsic differences between the Euclidean and wrapped GP models.
Note that we do not compare against VAE-based approaches~\citep{miolane:rvae:2020,hadi:rss:2021} as these models are structurally different from GPLVMs. For example, as discussed in Sec.~\ref{sec:intro}, they rely on handcrafted uncertainty quantification. 
Experimental settings, geodesics optimization, and training parameters for all models are detailed in App.~\ref{app:experimental_details}. A supplementary video and open source code can be found in \url{https://sites.google.com/view/riemann2}.\looseness=-1

\subsection{Illustrative Example on $\euclideanspace^2\times \sphere^2$}
\label{subsec:R2xS2}
\vspace{-0.2cm}
To illustrate the main differences between \riemannsq and the baselines, we first learn $2$-dimensional latent spaces out of trajectories 
on $\euclideanspace^2\times \sphere^2$. As in~\citep{hadi:rss:2021}, we consider a $\mathsf{J}$ shape in $\euclideanspace^2$ and a $\mathsf{C}$ shape projected on $\sphere^2$.
All approaches are augmented with a GPDM latent prior to account for the temporal structure of the trajectories.
Figure~\ref{fig:experiments:toy_experiment} shows the learned latent spaces as well as a geodesic in each latent space and the resulting decoded trajectory. Note that the GPLVM and pGPLVM share the same latent space as the pullback metric of the pGPLVM is obtained after training the GPLVM. This also applies to WGPLVM and \riemannsq. The background colors in the first and second columns represent the model uncertainty and the magnification factor $\sqrt{\det(\tilde{\bm{G}})}$ of \riemannsq and pGPLVM pullback metrics.\looseness=-1 

We observe that the magnification factor is low in the data manifold and high on its boundary, for both pullback approaches. This induces the Riemannian geodesic to travel along the data manifold, while Euclidean geodesics, i.e., straight lines, cross empty regions in the latent space. Consequently, the decoded Riemannian geodesics of pGPLVM and \riemannsq resemble the training trajectories unlike the decoded Euclidean geodesics of GPLVM and WGPLVM. However, only the decoded trajectories obtained from the manifold-aware WGPLVM and \riemannsq intrinsically stay on $\euclideanspace^2\times \sphere^2$. These observations are quantitatively supported by the metrics reported in Table~\ref{tab:experiments:all_experiment}, which reports the percentage of each geodesic belonging to $\euclideanspace^2\times \sphere^2$, and their dynamic time warping distance (DTWD) indicating their resemblance with the training data. Overall, \riemannsq is the only model that is manifold-aware and captures the underlying data manifold.\looseness=-1 

\begin{table}[tbp]
\caption{Percentage of geodesics on manifold ($\%$) and average dynamic time warping distance (DTWD) between the geodesic of each model and the demonstrations.} 
\label{tab:experiments:all_experiment}
\begin{center}
\vspace{-0.4cm}
\resizebox{\columnwidth}{!}{
\begin{tabular}{ll|cccc}
& &\textbf{GPLVM}&\textbf{pGPLVM}&\textbf{WGPLVM}&\riemannsq\\
\hline 
\multirow{2}{3em}{$\euclideanspace^2 \!\times\!\mathcal{S}^2 $} 
& $\%$ & $0.01$ & $0.0$ & $\bm{1.0}$ & $\bm{1.0}$ \\
& DTWD & $13.2\pm0.98$ & $\bm{3.4\pm0.77}$ & $11.9\pm0.91$ & $3.8\pm0.91$ \\
\hline
\multirow{2}{3em}{$\euclideanspace^3\!\times\!\mathcal{S}^3 $}& $\%$ & $0.02$ & $0.04$ & $\bm{1.0}$ & $\bm{1.0}$ \\
& DTWD & $1.08\pm0.06$ & $0.63\pm0.11$ & $0.94\pm0.09$ & $\bm{0.57\pm0.18}$ \\
\hline
$\SPD^{15}$ & $\%$ & $0.76$ & $\bm{1.0}$ & $\bm{1.0}$ & $\bm{1.0}$ \\
\end{tabular}
}
\vspace{-0.6cm}
\end{center}
\end{table}

\vspace{-0.1cm}
\subsection{Robot Motion Synthesis on $\euclideanspace^3\times \sphere^3$}
\vspace{-0.2cm}
Learning from demonstrations (LfD) consists of modeling and generating robot motions from a set of human examples. 
In this context, \citet{hadi:rss:2021} proposed to learn a latent Riemannian manifold from demonstrations and to leverage the corresponding geodesics to synthesize robot motions.
They considered a common LfD downstream task, namely a reach-and-pick-up task performed by a $7$-DoF robotic arm. The data was collected from real-world human demonstrations and consisted of end-effector positions and rotations encoded as points in $\euclideanspace^3\times \sphere^3$. 
We evaluate \riemannsq in a subset of this dataset, consisting of one starting and one target point. We here employ back-constrained models, where $k^{\manifold}$ is defined as the product $k^{\euclideanspace}\times k^{\sphere}$ of an Euclidean SE and sphere SE~\citep{Borovitskiy20:GPManifolds}, for the wrapped models. For Euclidean models, we use only an Euclidean SE kernel. 

\begin{figure}[tbp]
    \centering
    \begin{subfigure}[b]{\linewidth}
        \includegraphics[trim={0.0cm 1.5cm 0.0cm 1.5cm},clip,width=0.4\linewidth]{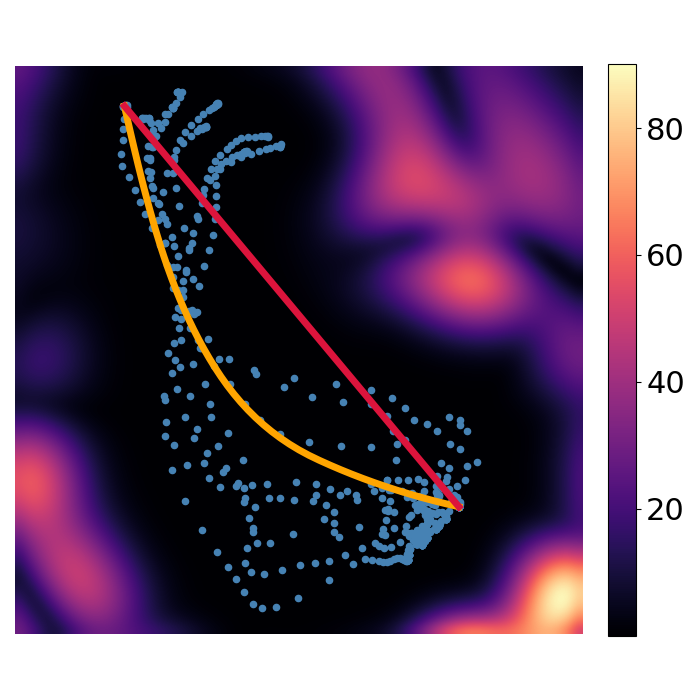}
        \includegraphics[trim={3.0cm 4.2cm 3.0cm 3.0cm},clip,width=0.4\linewidth]{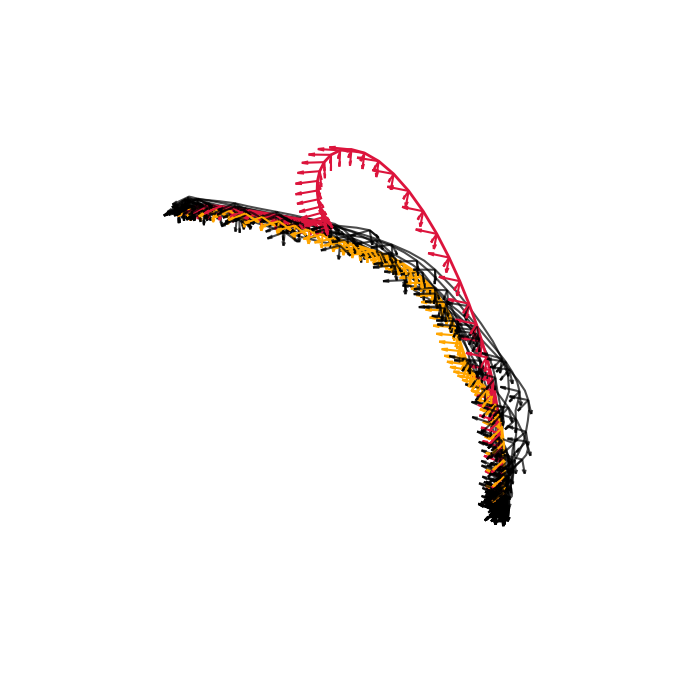}
        \caption{WPGLVM (\crimsonline) and \riemannsq (\yellowline).}
    \end{subfigure}
    
    \begin{subfigure}[b]{\linewidth}
        \includegraphics[trim={0.0cm 1.5cm 0.0cm 1.5cm},clip,width=0.4\linewidth]{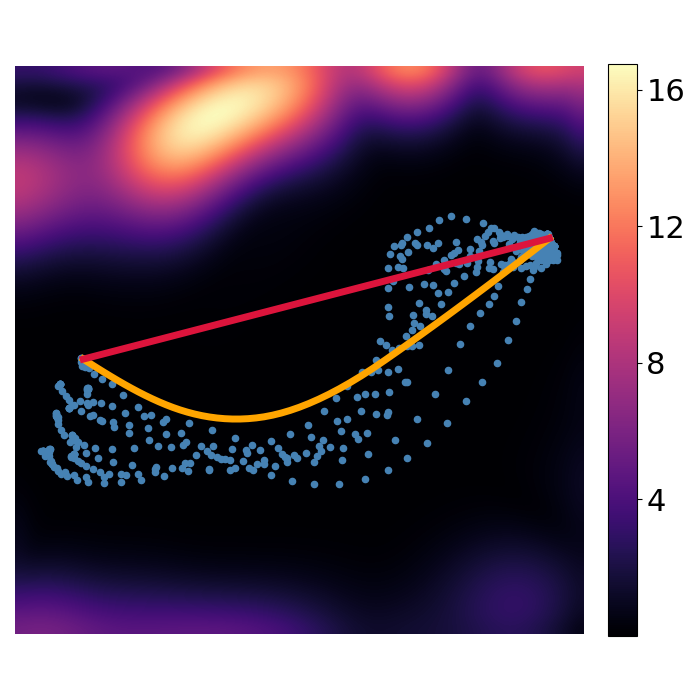}
        \includegraphics[trim={3.0cm 4.0cm 3.0cm 4.2cm},clip,width=0.4\linewidth]{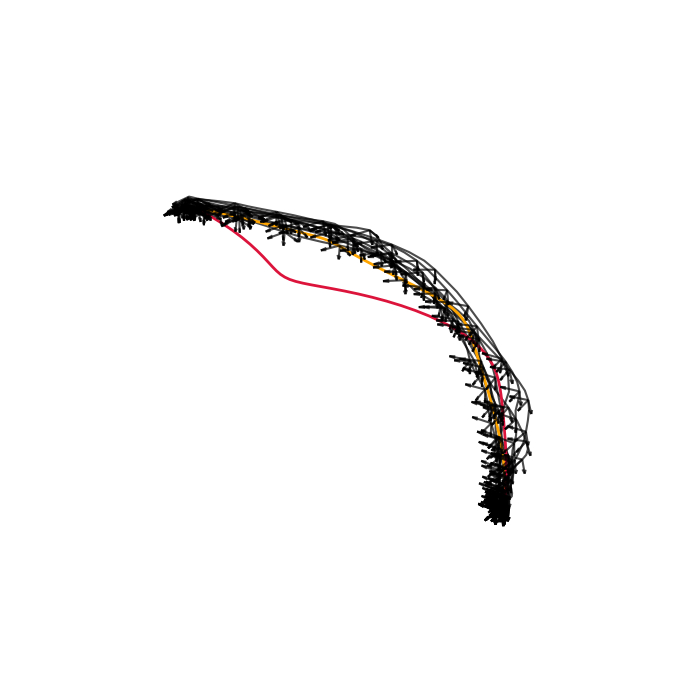}
        \caption{GPLVM (\crimsonline) and pGPLVM (\yellowline). \vspace{-0.2cm}}
        \label{fig:experiments:pickplace_experiment-gplvm}
    \end{subfigure}
    \caption{$\euclideanspace^3 \times \sphere^3$: \emph{Left}: Latent variables (\dodgerbluecircle) with magnification factor of the pullback metrics. One geodesic is depicted per model in the corresponding latent space. \emph{Right}: Demonstrations (\blackline) and reconstructions represented as positions and rotation frames. 
    Rotations are not depicted for GPLVM and pGPLVM as their reconstructions do not lie on $\sphere^3$.}
    \vspace{-0.3cm}
    \label{fig:experiments:pickplace_experiment}
\end{figure}

\begin{figure}[tbp]
    \centering
    \begin{subfigure}[b]{\linewidth}
        \includegraphics[trim={1.8cm 1.8cm 1.8cm 1.8cm},clip,width=0.24\linewidth]{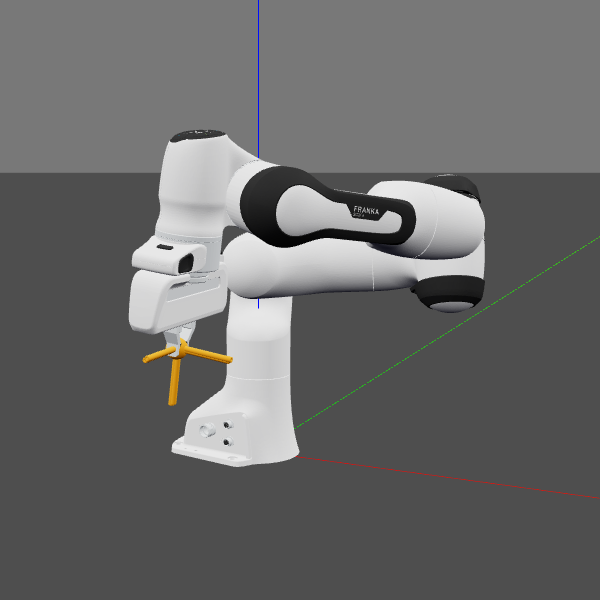}
        \includegraphics[trim={1.8cm 1.8cm 1.8cm 1.8cm},clip,width=0.24\linewidth]{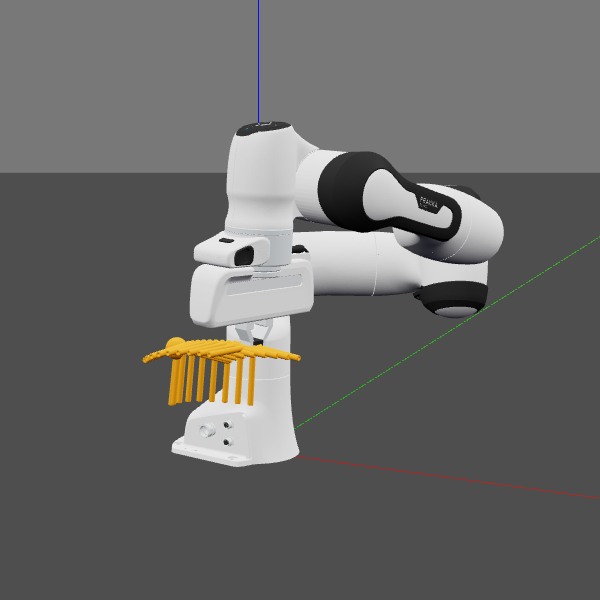}
        \includegraphics[trim={1.8cm 1.8cm 1.8cm 1.8cm},clip,width=0.24\linewidth]{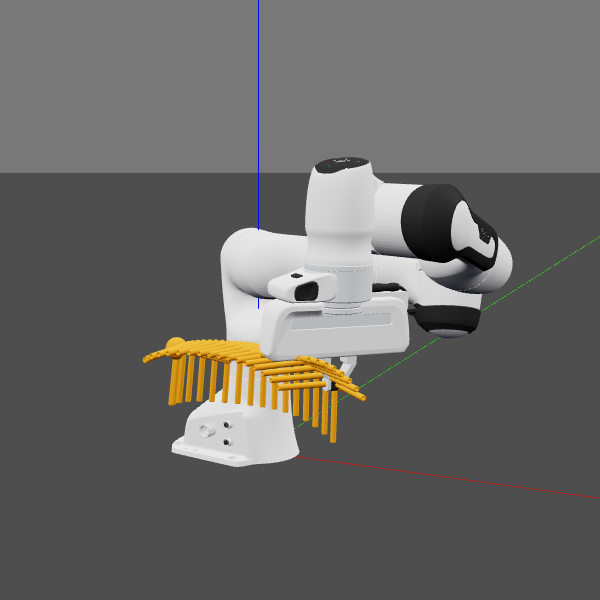}
        \includegraphics[trim={1.8cm 1.8cm 1.8cm 1.8cm},clip,width=0.24\linewidth]{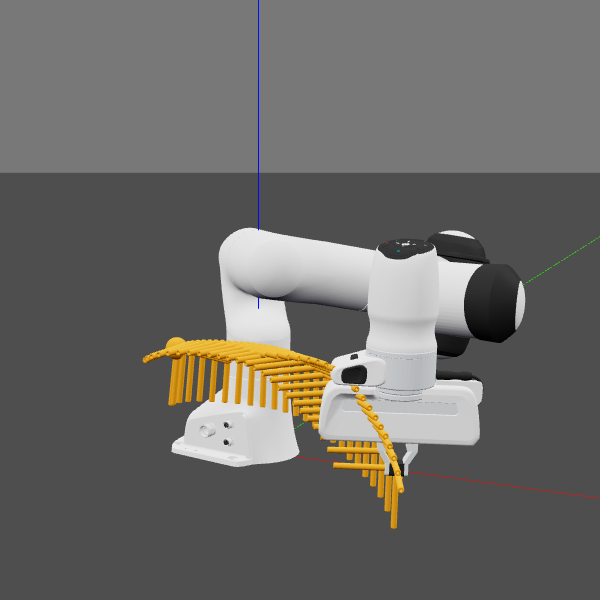}
        \caption{\riemannsq\!\!.}
    \end{subfigure}
    
    \begin{subfigure}[b]{\linewidth}
        \includegraphics[trim={1.8cm 1.8cm 1.8cm 1.8cm},clip,width=0.24\linewidth]{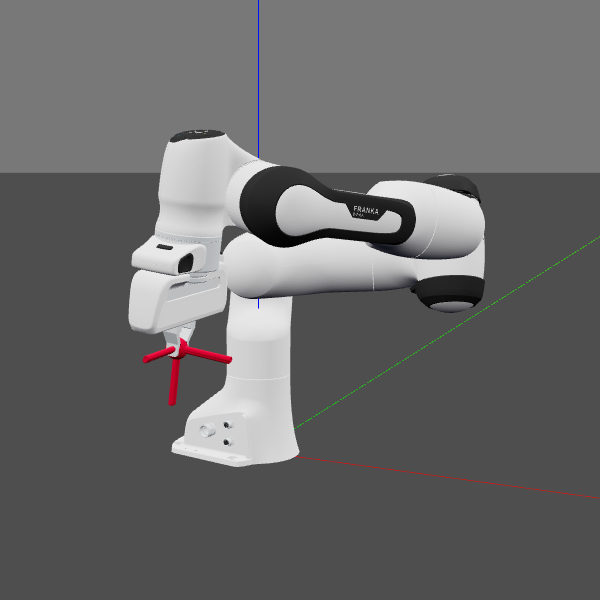}
        \includegraphics[trim={1.8cm 1.8cm 1.8cm 1.8cm},clip,width=0.24\linewidth]{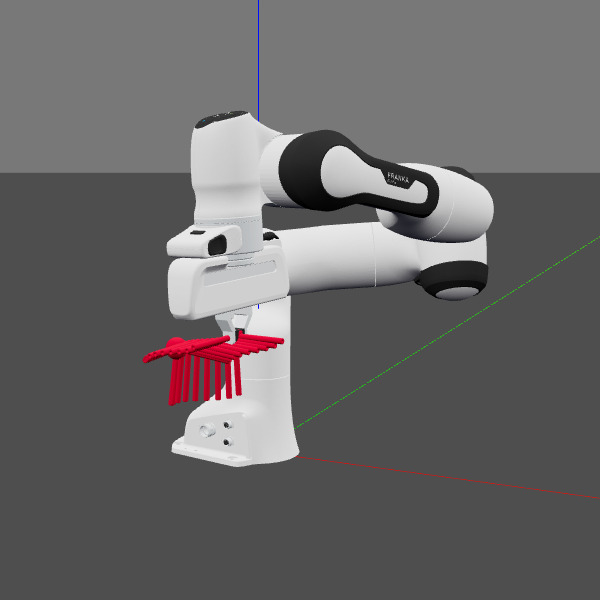}
        \includegraphics[trim={1.8cm 1.8cm 1.8cm 1.8cm},clip,width=0.24\linewidth]{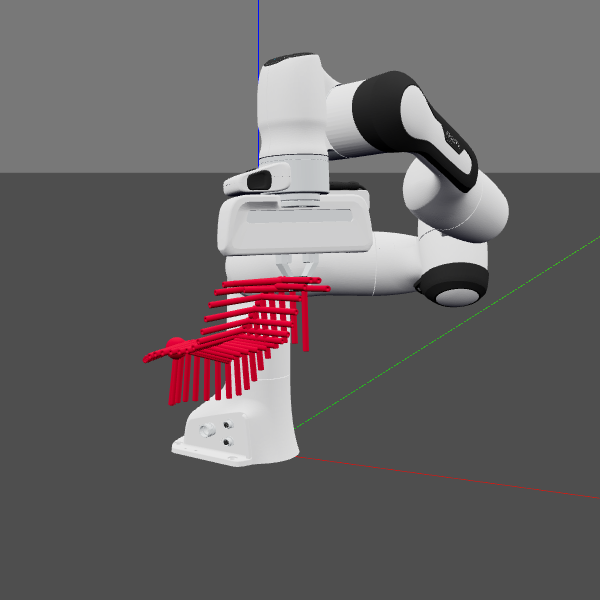}
        \includegraphics[trim={1.8cm 1.8cm 1.8cm 1.8cm},clip,width=0.24\linewidth]{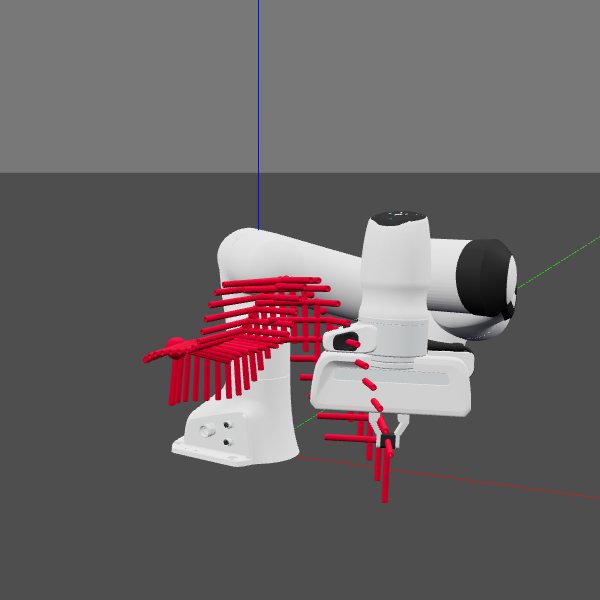}
        \caption{WGPLVM.\vspace{-0.2cm}}
    \end{subfigure}
    \caption{$\euclideanspace^3 \times \sphere^3$: Robot motions generated from the decoded WGPLVM and \riemannsq geodesics. End-effector trajectories are represented as position and rotation frames. 
    }
    \vspace{-0.4cm}
    \label{fig:experiments:pickplace_experiment_robots}
\end{figure}

Figure~\ref{fig:experiments:pickplace_experiment}-\emph{left} shows the latent spaces learned by the pullback models along with the corresponding magnification factors, as well as one geodesic for each model. The corresponding decoded trajectories are represented as frames on the \emph{right} panels. The pullback metrics of pGPLVM and \riemannsq encode the data manifold in the latent space and generate geodesics that follow the data distribution (see also the DTWD in Table~\ref{tab:experiments:all_experiment}). However, pGPLVM, as the GPLVM, fail to generate valid orientations in $\sphere^3$, as shown in Fig.~\ref{fig:experiments:pickplace_experiment-gplvm}-\emph{right} and Table~\ref{tab:experiments:all_experiment}. In contrast, \riemannsq generates valid trajectories resembling the training data in $\euclideanspace^3\times \sphere^3$. We use the decoded trajectories of WGPLVM and \riemannsq as reference positions and orientations to be tracked by the robot to execute the reach-and-pick task. The resulting robot motions are shown in Fig.~\ref{fig:experiments:pickplace_experiment_robots}. 
Note that GPLVM and pGPLVM cannot be used to generate robot motions as they do not lead to valid orientations.\looseness=-1

\vspace{-0.1cm}
\subsection{Manipulability Learning in $\euclideanspace^2 \times \SPD^2$}
\vspace{-0.2cm}
\label{sec:experiments:manipulability}
\begin{figure*}
    \centering
    \begin{subfigure}[b]{.95\textwidth}
    \centering
        \includegraphics[trim={0.0cm 1.5cm 0.0cm 1.5cm},clip,width=0.2\textwidth]{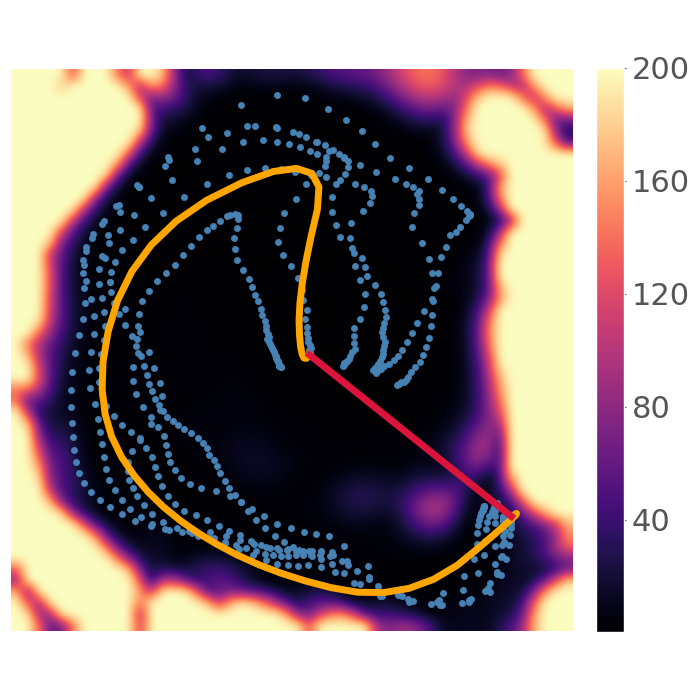}
        \includegraphics[trim={0.0cm 1.5cm 0.0cm 1.5cm},clip,width=0.2\textwidth]{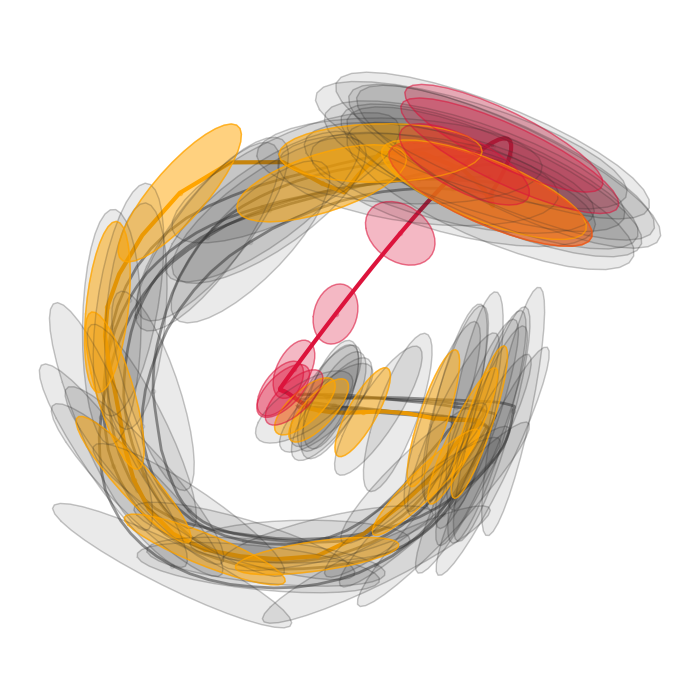}
        \includegraphics[trim={0.0cm 0.0cm 10.0cm 0.0cm},clip,width=0.15\textwidth]{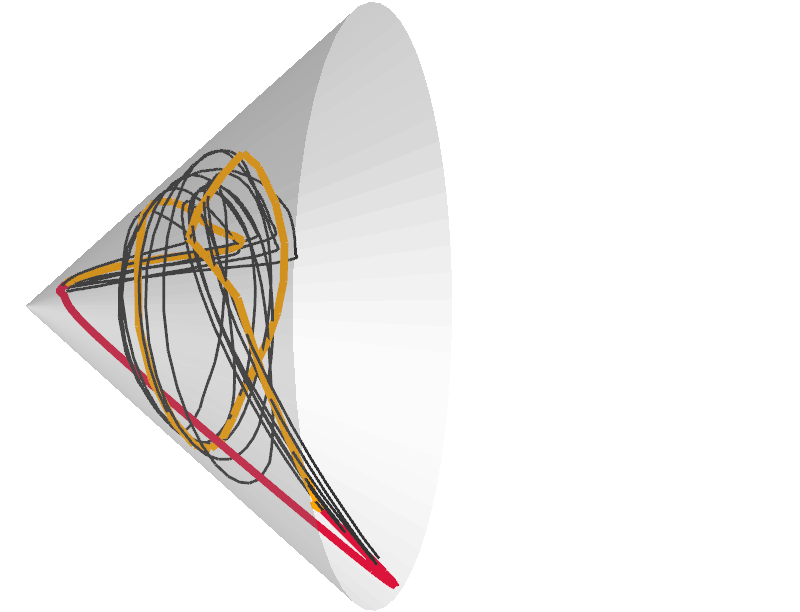}
        \includegraphics[trim={5.0cm 0.3cm 5.0cm 0.5cm},clip,width=0.3\textwidth]{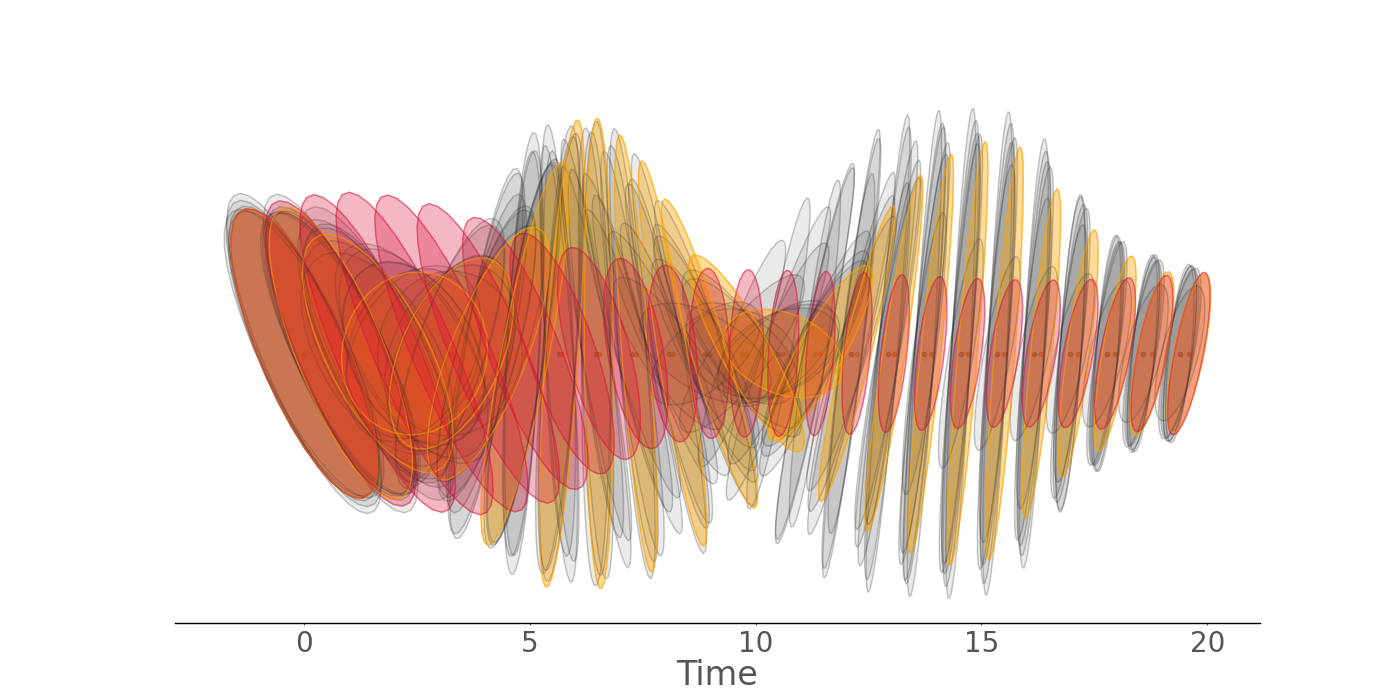}
        \caption{WPGLVM and \riemannsq latent space with one (decoded) geodesic each (\crimsonline, \yellowline). }
    \end{subfigure}%

    \begin{subfigure}[b]{.95\textwidth}
    \centering
        \includegraphics[trim={0.0cm 1.5cm 0.0cm 1.5cm},clip,width=0.2\textwidth]{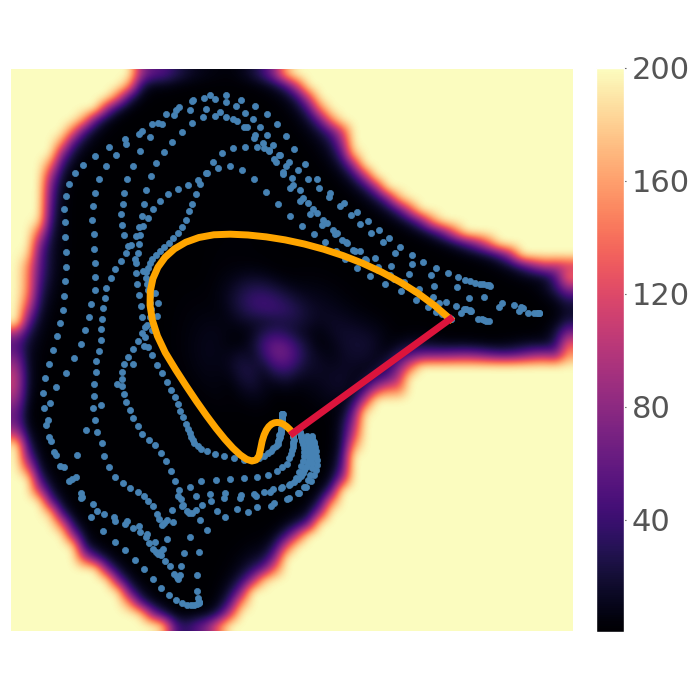}
        \includegraphics[trim={0.0cm 1.5cm 0.0cm 1.5cm},clip,width=0.2\textwidth]{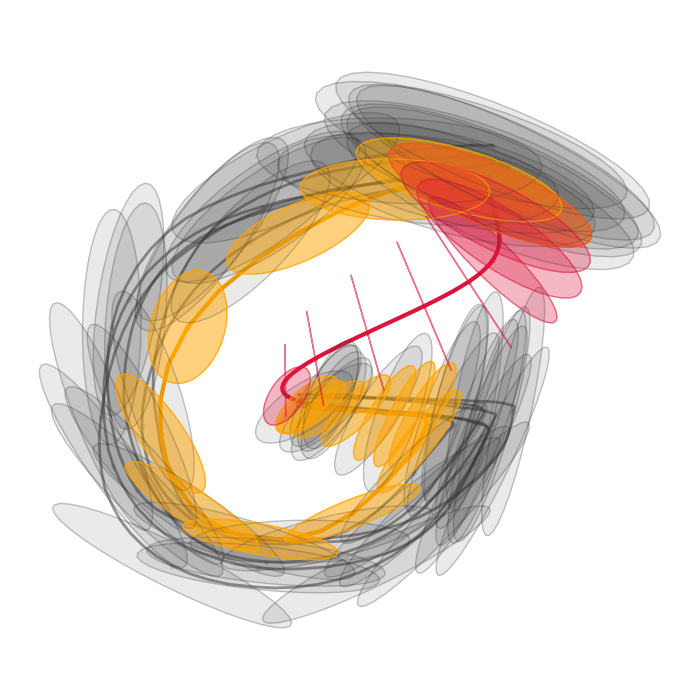}
        \includegraphics[trim={0.0cm 0.0cm 10.0cm 0.0cm},clip,width=0.15\textwidth]{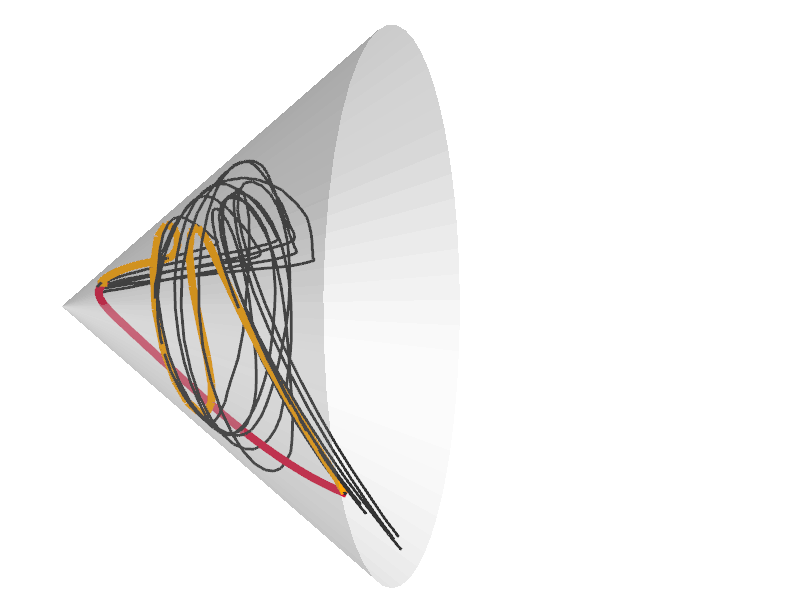}
        \includegraphics[trim={5.0cm 0.3cm 5.0cm 0.5cm},clip,width=0.3\textwidth]{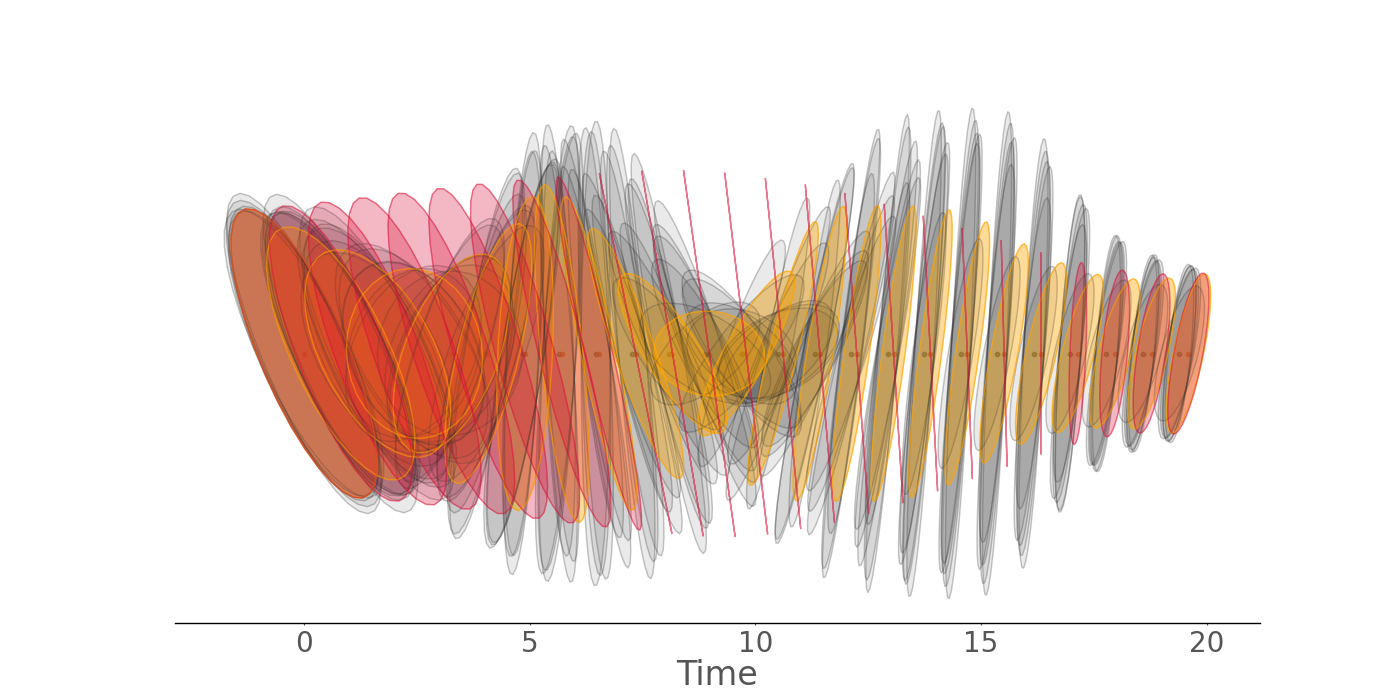}
        \caption{GPLVM and pGPLVM latent space with one (decoded) geodesic each (\crimsonline, \yellowline).
        \vspace{-0.2cm}}
    \end{subfigure}%
    \caption{$\euclideanspace^2 \times \SPD^2$: From \emph{left} to \emph{right}: Latent variables (\dodgerbluecircle) with magnification factor of the pullback metrics, demonstrations (\blackline,\grayellipse) and reconstructions depicted as curves and ellipsoids in $\euclideanspace^2$, on the manifold $\SPD^2$, and as ellipsoids over time. 
    }
    \vspace{-0.2cm}
    \label{fig:experiments:spd_experiment}
\end{figure*}

\begin{table*}[tbp]
\caption{Percentage of geodesics on manifold ($\%$) and average dynamic time warping distance (DTWD) between the geodesic and the demonstrations for the manipulability learning experiment on $\euclideanspace^2\times\SPD^2 $. } 
\label{tab:experiments:spd_experiment}
\begin{center}
\vspace{-0.4cm}
\resizebox{0.8\textwidth}{!}{
\begin{tabular}{l|cccc|cccc}
& \multicolumn{4}{c|}{\textbf{Independent output dimensions}} & \multicolumn{4}{c}{\textbf{Correlated output dimensions}} \\
&\textbf{GPLVM}&\textbf{pGPLVM}&\textbf{WGPLVM}&\riemannsq &\textbf{GPLVM}&\textbf{pGPLVM}&\textbf{WGPLVM}&\riemannsq \\
\hline 
$\%$ & $0.52$ & $\bm{1.0}$ & $\bm{1.0}$ & $\bm{1.0}$ & $\bm{1.0}$ & $\bm{1.0}$ & $\bm{1.0}$ & $\bm{1.0}$ \\
DTWD & $18.3\pm1.8$ & $6.1\pm1.9$ & $18.3\pm 1.8$& $3.5\pm 0.8$ & $17.0 \pm 1.9$ & $4.9\pm 1.7$ &$16.3\pm1.9$ & $\bm{3.5\pm0.3}$ \\
\end{tabular}
}
\vspace{-0.5cm}
\end{center}
\end{table*}

Next, we test \riemannsq on a challenging robot learning downstream task: Manipulability learning on a highly-redundant planar robot. Manipulability ellipsoids are defined as functions of the robot joint configuration $\bm{q}$ as $\bm{M}(\bm{q}) = \bm{J}(\bm{q})^\top \bm{J}(\bm{q})$, with $\bm{J}(\bm{q})$ being the robot Jacobian\footnote{$\bm{J}(\bm{q})$ is computed from the robot kinematics and should not be confused with the Jacobian introduced in Sec.~\ref{sec:background:riemannian}.}. They indicate the preferred directions of velocity control.
The considered task involves learning jointly a robot end-effector trajectory and a manipulability profile. Robot motions are then generated by following the desired end-effector trajectory as the main control task, and the desired manipulability profile as a secondary objective, as proposed by~\citep{jaquier:manipulability:2021}. 
In this setting, we learn a Riemannian submanifold from data in $\euclideanspace^2\times\SPD^2$. Specifically, we consider training data consisting of the position trajectories and manipulability profiles recorded from a simulated highly-redundant planar robot following a $\mathsf{G}$-shape trajectory on a 2D plane. In the demonstrations, the robot velocity manipulability is aligned with the direction of motion, thereby allowing it to follow a given trajectory at a higher speed. Note that this is a desired feature in robotic tasks, as this allows a robot to shape its posture according to the task requirements~\citep{jaquier:manipulability:2021}.
All models are augmented with a GPDM prior and with back constraints. The kernel $k^{\manifold}$ is defined as the product $k^{\euclideanspace}\times k^{\SPD}$ of a Euclidean SE and SPD SE~\citep{Azangulov23:NonCompactKernels} kernels, and as a Euclidean SE kernel, for the wrapped and Euclidean models, respectively.\looseness-1

Figure~\ref{fig:experiments:spd_experiment} shows the learned latent spaces, alongside magnification factors for the pullback metrics, geodesic trajectories, and decoded position and manipulability profiles. As for the previous experiments, the decoded geodesics generated by pGPLVM and \riemannsq resemble the data distribution due to the pullback metric. Unlike the experiments with data on the hypersphere manifold, the trajectories generated by pGPLVM fulfill the SPD manifold constraints (see also Table~\ref{tab:experiments:spd_experiment}). This is due to the geometry of the SPD manifold (see third panels of Fig.~\ref{fig:experiments:spd_experiment}), for which linear interpolation between nearby SPD matrices remains in the SPD cone. However, the trajectories generated by \riemannsq match the training data more closely than those obtained by pGPLVM, as shown by the DTWD values in Table~\ref{tab:experiments:spd_experiment}. Moreover, crossing empty regions, e.g., with the GPLVM geodesic, still results in invalid trajectories.

Next, we analyse the importance of considering correlated output dimensions within the multitask models. Table~\ref{tab:experiments:spd_experiment} shows that models incorporating correlated output dimensions consistently generate trajectories that more closely align with the training data patterns than those assuming independent dimensions. \riemannsq with correlated outputs achieves the best results over all, showing the relevance of accounting for the data and latent space geometries, as well as the correlations between position and manipulability.
Note that \riemannsq with independent outputs achieves a similar DTWD mean, but higher variance. In contrast, the manipulability profiles are better reconstructed with correlated outputs, as illustrated in Fig.~\ref{fig:experiments:spd_supplementary} of App.~\ref{app:additional_results}.

\subsection{Brain Connectomes in $\SPD^{15}$}
\vspace{-0.2cm}
Finally, we test the ability of \riemannsq to learn a submanifold of high-dimensional data. Similarly as~\citep{miolane:rvae:2020}, we consider resting-state functional brain connectome data from the ``1200 Subjects release'' of the Human Connectome Project~\citep{VanEssen13:HumanConnectome}. We learn a submanifold from brain connectomes from $200$ subjects out of the $812$ release. Each connectome is represented as a point in $\SPD^{15}$.

\begin{figure}
    \centering
    \begin{subfigure}[b]{.49\linewidth}
        \includegraphics[trim={0.0cm 1.5cm 0.0cm 1.2cm},clip,width=0.93\linewidth]{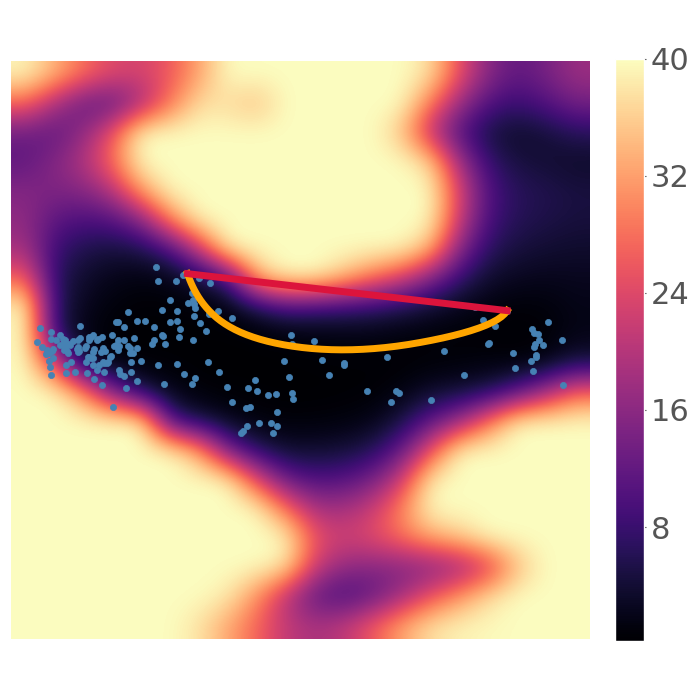}
        \caption{WPGLVM (\crimsonline) and \\ \riemannsq (\yellowline).}
    \end{subfigure}
    \begin{subfigure}[b]{.49\linewidth}
        \includegraphics[trim={0.0cm 1.8cm 0.0cm 1.3cm},clip,width=\linewidth]{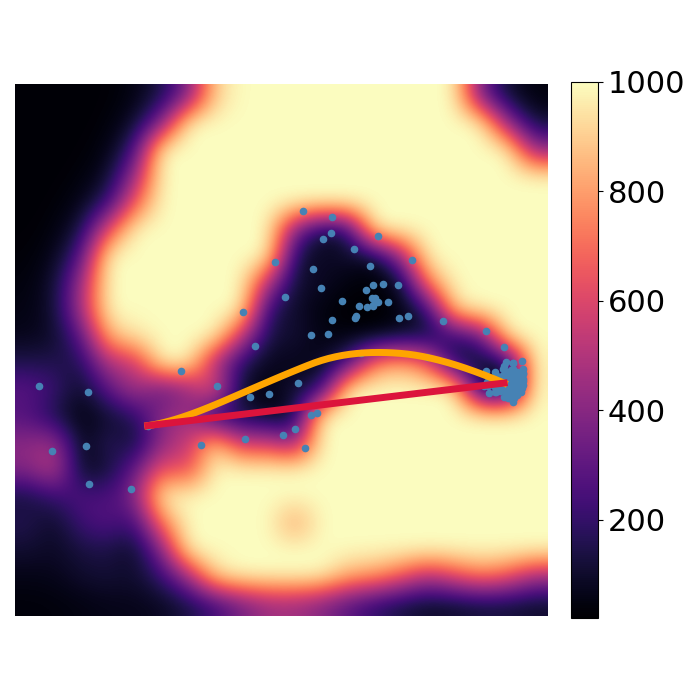}
        \caption{GPLVM (\crimsonline) and \\ pGPLVM (\yellowline).}
    \end{subfigure}
    \caption{$\SPD^{15}$: Latent variables (\dodgerbluecircle) with magnification factor of the pullback metrics. One Euclidean (\crimsonline) and one Riemannian (\yellowline) geodesic are depicted for each model.
    }
    \vspace{-0.4cm}
    \label{fig:experiments:connectomes}
\end{figure}

Figure~\ref{fig:experiments:connectomes} shows the latent spaces learned with GPLVM, pGPLVM, WGPLVM, and \riemannsq, alongside magnification factors for the pullback metrics and geodesic trajectories. We observe that most of the latent variables are cluttered in the middle-right part of the GPLVM and pGPLVM latent space, while they are more evenly distributed in the case of \riemannsq and WPGLVM. Moreover, the pGPLVM magnification factor does not draw a clear boundary around the data manifold in this case. As for previous experiments, the Euclidean models do not guarantee that the decoded geodesics belong to the given manifold (see Table~\ref{tab:experiments:all_experiment}).

\vspace{-0.1cm}
\subsection{Comparison with Metric Learning}
\vspace{-0.2cm}

\begin{figure}
    \centering
    \includegraphics[trim={.0cm 1.5cm 0.0cm 1.7cm},clip,width=0.45\linewidth]{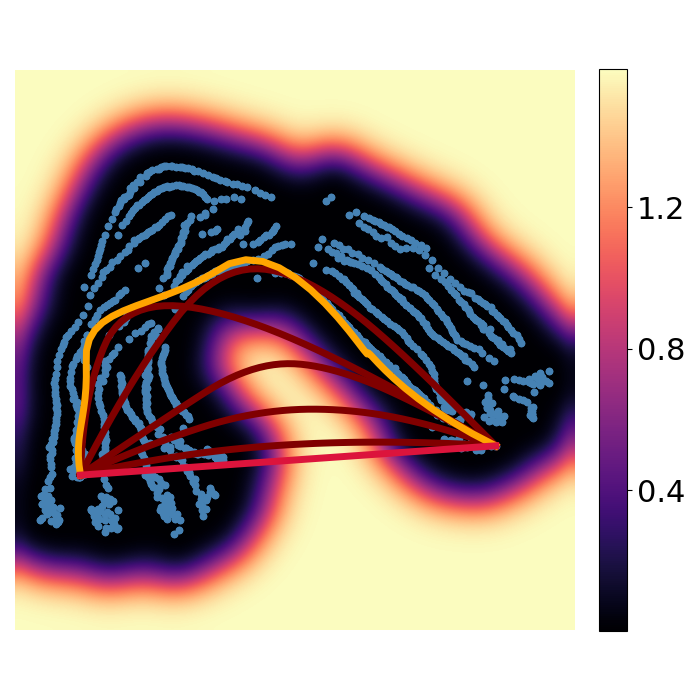}
    \includegraphics[trim={0.0cm 1.5cm 0.0cm 1.7cm},clip,width=0.45\linewidth]{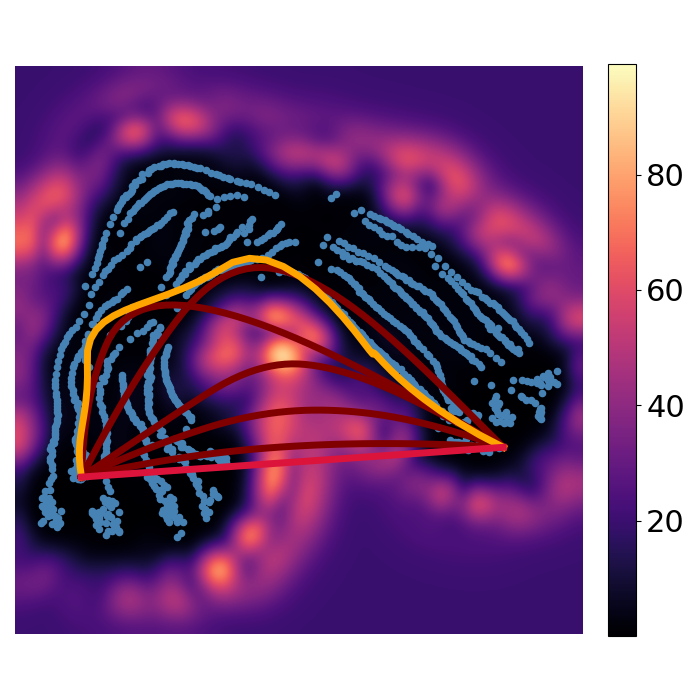}
    \caption{Metric learning in $\euclideanspace^2 \times \sphere^2$: From \emph{left} to \emph{right}: Latent variables (\dodgerbluecircle) with model uncertainty and magnification factor of the pullback metrics with one WGPLVM (\crimsonline), one \riemannsq (\yellowline), and several metric-learning (\maroonline) geodesics for different parameters $\sigma$.}
    \vspace{-0.3cm}
    \label{fig:MetricLearning}
\end{figure}

We compare \riemannsq with metric learning in the latent space of a trained WGPLVM. 
Following~\citep{Lebanon03:LearningRiemannianMetric}, we use a metric $(\det{\bm{G}(\bm{x})})^{-1/2} \propto p(\bm{x})$, where $p(\bm{x})$ is a density estimated from data. 
As the density is a scalar measure, we chose the metric to be isotropic, i.e., $\bm{G}(\bm{x}) = \lambda(\bm{x}) \bm{I}$ with $\lambda^{-Q/2}(\bm{x}) \propto p(\bm{x})$. 
We define $p(\bm{x})$ as a kernel density estimate of the latent variables, i.e., $p(\bm{x}) = \sum_{n=1}^N \frac{1}{(2\pi)^{Q/2} \sigma^D} \exp(- \frac{d(\bm{x}, \bm{x}_n)}{2\sigma^2})$, where $\bm{x}_n$ are the training latent variables and $\sigma$ is a hyperparameter defining the density variance. We compute Riemannian metrics $\bm{G}(\bm{x})$ derived from the kernel density estimate on the illustrative example of Sec.~\ref{subsec:R2xS2} for different values of $\sigma$, as well as corresponding geodesics for each metric with the same start and end points as in Fig.~\ref{fig:experiments:toy_experiment}. The corresponding results are reported in Fig.~\ref{fig:MetricLearning} and Table~\ref{tab:experiments:metric_learning_experiment}. We observe that the WGPLVM ensures that all decoded geodesics stay on $\mathbb{R}^2 \times \mathbb{S}^2$. However, the aforementioned metric learning approach is sensitive to the hyperparameter $\sigma$: Its DTWD is similar to that of \riemannsq for $\sigma=0.25$, indicating the similarity of the generated geodesic with the training data. However, the DTWD drastically increases for other hyperparameter values, leading to trajectories that strongly differ from the training data, similar to the WGPLVM geodesic. 

Compared to metric learning on the latent space, pullback metrics offer several advantages: They avoid introducing additional parameters, thus simplifying the model,and they eliminate the need for a separate learning step after training the latent variables, streamlining the overall training process. Furthermore, metric learning is very sensitive to parameter choices.

\begin{table}[tbp]
\caption{Percentage of geodesics on manifold ($\%$) and average DTWD between the geodesic of each metric-learning model and the demonstrations.} 
\label{tab:experiments:metric_learning_experiment}
\begin{center}
\vspace{-0.4cm}
\resizebox{\columnwidth}{!}{
\begin{tabular}{l|ccccc} 
$\sigma$ & $0.1$ & $0.25$ & $0.5$ & $1.0$ & $2.0$ \\
\hline
$\%$ & $\bm{1.0}$ &$\bm{1.0}$ &$\bm{1.0}$ & $\bm{1.0}$ & $\bm{1.0}$ \\
DTWD & $7.3\pm 0.49$ & $\bm{3.6\pm0.80}$ & $11.6\pm0.65$ & $13.1\pm0.71$ & $12.3 \pm 0.81$\\
\end{tabular}
}
\vspace{-0.6cm}
\end{center}
\end{table}


\vspace{-0.1cm}
\section{Conclusions}
\vspace{-0.2cm}
Our goal of learning a Riemannian submanifold from Riemannian data was twofold: (1) When learning latent representations, any operation on the latent space such distances or geodesics between two embeddings, must comply with the underlying data manifold; (2) We must guarantee that the decoded latent variables lie on the Riemannian manifold of interest. Both objectives proved to be notably important when geodesics were employed as a motion generation mechanism, which may unlock applications on avatars animation or humanoid robots control. However, as \riemannsq generalizes previous works, its features can also be separately employed to learn geometry-aware GP generative models, or to analyze the geometry of GPLVMs, for single or multitask GPs. For instance, if adhering to the ambient geometry is not critical, then learning a Riemannian pullback metric may suffice for data manifold estimation and sample generation.
Similarly as GPLVMs, it is often relevant to incorporate additional smoothness constraints on the latent spaces of \riemannsq through priors and back constraints. The latter increases the computational cost of the model due to the use of Riemannian kernels. However, they are necessary to account for the data geometry.\looseness-1

\clearpage
\subsubsection*{Acknowledgements}
N. J. was partially supported by the Wallenberg AI, Autonomous Systems and Software
Program (WASP) funded by the Knut and Alice Wallenberg Foundation.
S. H. was supported by a research grant (42062) from VILLUM FONDEN, and also partly funded by the Novo Nordisk Foundation through the Center for Basic Research in Life Science (NNF20OC0062606).
S. H. received funding from the European Research Council (ERC) under the European Union's Horizon Programme (grant agreement 101125003).

\bibliographystyle{abbrvnat}
\bibliography{biblio}

\clearpage
\section*{Checklist}



 \begin{enumerate}

 \item For all models and algorithms presented, check if you include:
 \begin{enumerate}
   \item A clear description of the mathematical setting, assumptions, algorithm, and/or model. Yes.
   \item An analysis of the properties and complexity (time, space, sample size) of any algorithm. Yes.
   \item (Optional) Anonymized source code, with specification of all dependencies, including external libraries. No, but the code will be released upon acceptance of the paper.
 \end{enumerate}

 \item For any theoretical claim, check if you include:
 \begin{enumerate}
   \item Statements of the full set of assumptions of all theoretical results. Yes.
   \item Complete proofs of all theoretical results. Yes.
   \item Clear explanations of any assumptions. Yes.     
 \end{enumerate}

 \item For all figures and tables that present empirical results, check if you include:
 \begin{enumerate}
   \item The code, data, and instructions needed to reproduce the main experimental results (either in the supplemental material or as a URL). No, the code will be released upon acceptance of the paper.
   \item All the training details (e.g., data splits, hyperparameters, how they were chosen). Yes.
         \item A clear definition of the specific measure or statistics and error bars (e.g., with respect to the random seed after running experiments multiple times). Yes.
         \item A description of the computing infrastructure used. (e.g., type of GPUs, internal cluster, or cloud provider). Not Applicable.
 \end{enumerate}

 \item If you are using existing assets (e.g., code, data, models) or curating/releasing new assets, check if you include:
 \begin{enumerate}
   \item Citations of the creator If your work uses existing assets. Yes.
   \item The license information of the assets, if applicable. Yes.
   \item New assets either in the supplemental material or as a URL, if applicable. Not Applicable.
   \item Information about consent from data providers/curators. Not Applicable.
   \item Discussion of sensible content if applicable, e.g., personally identifiable information or offensive content. Not Applicable.
 \end{enumerate}

 \item If you used crowdsourcing or conducted research with human subjects, check if you include:
 \begin{enumerate}
   \item The full text of instructions given to participants and screenshots. Not Applicable.
   \item Descriptions of potential participant risks, with links to Institutional Review Board (IRB) approvals if applicable. Not Applicable.
   \item The estimated hourly wage paid to participants and the total amount spent on participant compensation. Not Applicable.
 \end{enumerate}

 \end{enumerate}

\appendix
\thispagestyle{empty}
\onecolumn
\section{Distribution of the pullback metric of \riemannsq}
\label{app:PullbackDistribution}
This section details the derivation of the pullback Riemannian metric of the latent space of \riemannsq, presented in Section~\ref{sec:riemannsq} of the main paper.

\subsection{The pullback metric}
\label{app:Pullback}
In this paper, we model manifold-valued functions $f = \Exp_{b(\cdot)}\circ\, \feuc: \euclideanspace^Q\to (\manifold, g)$ using WGPs. We then use this mapping to pull back the metric $g$, defining a custom geometry in the latent space $\euclideanspace^d$. Given two tangent vectors $\bm{v}_1, \bm{v}_2 \in \mathcal{T}_{\bm{x}}\euclideanspace^d$, the pullback metric $\tilde{g} = f^* g$, with $^*$ denoting the pullback operator, is defined as in Eq.~\eqref{eq:background:pullback_definition} \citep[Chap. 2]{Lee18Riemann} by,
\begin{equation*}
    \label{eq:appendix:pullback_metric}
    \tilde{g}_{\bm{x}} (\bm{v}_1, \bm{v}_2) = g_{f(\bm{x})}\Big(\dd f_{\bm{x}}(\bm{v}_1), \dd f_{\bm{x}}(\bm{v}_2)\Big).
\end{equation*}
In other words, we carry the tangent vectors from $\mathcal{T}_{\bm{x}}\euclideanspace^d$ to $\mathcal{T}_{f(\bm{x})}\manifold$ using the differential of $f$, and compute their inner product there. 
Given a choice of coordinates, the pullback metric $\tilde{g}$ can be represented in matrix form. Specifically, considering the matrix representation of $(f^* g)_{\bm{x}}$, $g_{f(\bm{x})}$ and $(\mathrm{d}f)_{\bm{x}}$, we get,
\begin{equation*}
    \bm{v}_1^\top [\tilde{g}_{\bm{x}}]\bm{v}_2 = \bm{v}_1^\top \tilde{\bm{G}}(\bm{x}) \bm{v}_2 = \bm{v}_1^\top \bm{J}_{f(\bm{x})}^\top \bm{G}(f(\bm{x})) \bm{J}_{f(\bm{x})} \bm{v}_2 = \bm{v}_1^\top \bm{J}_{f(\bm{x})}^\top [g_{f(\bm{x})}] \bm{J}_{f(\bm{x})} \bm{v}_2.
\end{equation*}
This implies that, in coordinates, $\tilde{\bm{G}} = \bm{J}_f^\top \bm{G} \bm{J}_f$. 
When the mapping is a WGP, all the components of this product are stochastic and induce a distribution over the metric $\tilde{\bm{G}}$. As~\citet{Tosi:UAI:2014}, we consider the expected pullback metric $\mathbb{E}[\tilde{\bm{G}}]$ as a point estimate of the pullback metric distribution, as explained in Section~\ref{sec:method:pullback_metric}. To obtain the expected pullback metric, we need to compute the Jacobian $\bm{J}_{f}=\bm{J}_{\Exp_{b(\cdot)}} \bm{J}_{\feuc}$. The computation of the first Jacobian $\bm{J}_{\Exp_{b(\cdot)}}$ is explained in Section~\ref{sec:method:pullback_metric}. The computation of the second Jacobian is briefly presented in Section~\ref{sec:method:jacobian_of_multitask_gp} and further elaborated next.

\subsection{The distribution of the Jacobian of a multitask Gaussian Process}
\label{sec:appendix:distribution_of_multitask_gp}
Here, we focus on computing the distribution of the Jacobian $\bm{J}_{\feuc}$ of a multitask GP $\feuc$ at a new test point $\bm{x}_*$ and elaborate the derivation presented in Section~\ref{sec:method:jacobian_of_multitask_gp}. First, we derive the joint distribution of the data and the transpose Jacobian given by Eq.~\eqref{eq:method:joint_distribution_of_data_and_jacobian}. 
Due to the linearity of the differentiation operator, the derivative of a GP $\feuc\sim \GP(\mu(\bm{x}), k(\bm{x},\bm{x}))$ is given by another GP~\citep[Ch. 9]{Rasmussen},
\begin{equation}
    \bm{J}_{\feuc(\bm{x})} = \frac{\partial \feuc (\bm{x})}{\partial \bm{x}}\sim \GP\left(\frac{\partial \mu(\bm{x})}{\partial \bm{x}}, \frac{\partial^2 k(\bm{x},\bm{x})}{\partial \bm{x}^2}\right).
\end{equation}
Therefore, the distribution of the Jacobian of $\feuc$ is governed by the partial derivatives of the kernel,
\begin{align}
    \cov\left(f_{j}(\bm{x}_i), \pderiv{f_m(\bm{x}_*)}{x_{r}}\right) &= \pderiv{k_{jm}(\bm{x}_i, \bm{x}_*)}{x^{(r)}}, \label{eq:appendix:cov_partial} \\ 
    \cov\left(\pderiv{f_{j}(\bm{x}_*)}{x^{(r)}}, \pderiv{f_m(\bm{x}_*)}{x^{(s)}}\right) &= \ppderiv{k_{jm}(\bm{x}_i, \bm{x}_*)}{x^{(r)}}{x^{(s)}}, \label{eq:appendix:cov_partial_partial}
\end{align}
where $f_j(\bm{x}_i)$ is the $j$-th component of $f(\bm{x}_i)$, and $x^{(r)}$ is the $r$-th component of a latent variable $\bm{x}$.

To compute the posterior distribution of the Jacobian $\bm{J}_{\feuc}$ at a test point $\bm{x}_*$, we consider the matrix $\bm{Y}$ containing our training outputs and the transpose Jacobian of $\feuc$ denoted $\bm{J}_{\feuc}^\top$,
\begin{equation*}
    \bm{Y} = \begin{bmatrix}
    y_1(\bm{x}_1) & y_2(\bm{x}_1) & \cdots & y_M(\bm{x}_1) \\
    y_1(\bm{x}_2) & y_2(\bm{x}_2) & \cdots & y_M(\bm{x}_2) \\
    \vdots & \vdots & \ddots & \vdots \\
    y_1(\bm{x}_N) & y_2(\bm{x}_N) & \cdots & y_M(\bm{x}_N) \\
    \end{bmatrix}_{N\times M}, \quad \text{and} \quad \bm{J}_{\feuc(\bm{x}_*)}^\top = \begin{bmatrix}
    \pderiv{y_1(\bm{x}_*)}{x^{(1)}} & \pderiv{y_2(\bm{x}_*)}{x^{(1)}} & \cdots & \pderiv{y_M(\bm{x}_*)}{x^{(1)}} \\
    \pderiv{y_1(\bm{x}_*)}{x^{(2)}} & \pderiv{y_2(\bm{x}_*)}{x^{(2)}} & \cdots & \pderiv{y_M(\bm{x}_*)}{x^{(2)}} \\
    \vdots & \vdots & \ddots & \vdots \\
    \pderiv{y_1(\bm{x}_*)}{x^{(Q)}} & \pderiv{y_2(\bm{x}_*)}{x^{(Q)}} & \cdots & \pderiv{y_M(\bm{x}_*)}{x^{(Q)}} \\
    \end{bmatrix}_{Q\times M}.
\end{equation*}
Using Eqs.~\eqref{eq:appendix:cov_partial} and~\eqref{eq:appendix:cov_partial_partial}), we first obtain the joint distribution of the data $\bm{Y}$ and transpose Jacobian $\bm{J}_{\feuc(\bm{x}_*)}^\top$ given by,
\begin{equation}
\label{eq:appendix:joint_distribution_of_data_and_jacobian}
    \begin{bmatrix}
    \vvec(\bm{Y}) \\
    \vvec(\bm{J}_{\feuc(\bm{x}_*)}^\top)
    \end{bmatrix} \sim \mathcal{N}\left(\bm{0}, \begin{bmatrix}
    \bm{K} & \partial \bm{K} \\
    \partial \bm{K}^\top & \partial^2 \bm{K} 
    \end{bmatrix}\right),
\end{equation}
where $\bm{K}$ is a block of size $NM\times MN$ which contains the covariances between the elements in $\vvec(\bm{Y})$, $\partial \bm{K}$ is a block of size $NM\times QM$ of partial derivatives w.r.t one latent dimension~\eqref{eq:appendix:cov_partial}, and $\partial^2 \bm{K}$ is a block of size $QM\times QM$ of second partial derivatives~\eqref{eq:appendix:cov_partial_partial}.

Previous literature on latent geometries in GPLVMs~\citep{Tosi:UAI:2014} modelled multivariate outputs using an independent GP for each dimension, or in other words, considered the output dimensions to be uncorrelated. In contrast, we model multivariate outputs using multitask GPs~\citep{Bonilla:multitaskGP:2007} via the multitask covariance of Eq.~\eqref{eq:background:multitask-gp-kernel}. This covariance is written succinctly as $k = k^{\tasks}\otimes k^{\bm{x}}$, 
where $k^{\tasks}$ is a symmetric positive-definite task covariance parametrized as $k^{\tasks} = \bm{B}\bm{B}^\top + \diag(\bm{v})$ with a (low-rank) matrix $\bm{B}$ and a diagonal vector of variances $\bm{v}$, and
$\otimes$ denotes the Kronecker product. For a multitask GP with covariance $k$, the kernel and partial derivatives of the joint distributions~\eqref{eq:appendix:joint_distribution_of_data_and_jacobian} are given by,
\begin{align}
    &\bm{K} = k^{\tasks}\otimes \bm{K}^{\bm{x}}, \quad\text{ with } \bm{K}^{\bm{x}} = \left[k^{\bm{x}}(\bm{x}_n, \bm{x}_{a})\right]_{n, a = 1}^{N}, \label{eq:appendix:gram_matrix} \\
    &\partial \bm{K} = k^{\tasks}\otimes \partial \bm{K}^{\bm{x}}, \quad\text{ with } \partial \bm{K}^{\bm{x}} = \left[\pderiv{k^{\bm{x}}(\bm{x}_n, \bm{x}_*)}{x^{r}}\right]_{n, r = 1}^{N,d}, \label{eq:appendix:partial_deriv_matrix} \\
    &\partial^2 \bm{K} = k^{\tasks}\otimes \partial^2 \bm{K}^{\bm{x}}, \quad\text{ with } \partial^2 \bm{K}^{\bm{x}} = \left[\ppderiv{k^{\bm{x}}(\bm{x}_*, \bm{x}_*)}{x^{r}}{x^{s}}\right]_{r, s = 1}^{d}. \label{eq:appendix:partial_partial_deriv_matrix}
\end{align}
The explicit computation of Eqs.~\eqref{eq:appendix:gram_matrix}-\eqref{eq:appendix:partial_partial_deriv_matrix} is provided in App.~\ref{sec:appendix:matrix_computations_of_multitask_kernel_derivatives}. We argue that these equations are intuitive as the derivative of a Kronecker product follows the product rule, and the tasks kernel does not depend on $\bm{x}$.

Second, we compute the posterior over $\vvec(\bm{J}_{\feuc(\bm{x}_*)}^\top)$ by conditioning the joint distribution~\eqref{eq:appendix:joint_distribution_of_data_and_jacobian} on $\vvec(\bm{Y})$. Using the multitask kernel expressions~\eqref{eq:appendix:gram_matrix}-\eqref{eq:appendix:partial_partial_deriv_matrix}, we have,
\begin{align}
    \vvec(\bm{J}_{\feuc(\bm{x}_*)}^\top) &\sim \mathcal{N}(\partial \bm{K}^\top \bm{K}^{-1}\vvec(\bm{Y}), \partial^2 \bm{K} - \partial \bm{K}^\top \bm{K}^{-1}\partial \bm{K}) \\
    &= \mathcal{N}\left((k^{\tasks\top}\otimes\partial \bm{K}^{\bm{x}\top}) ((k^{\tasks})^{-1}\otimes (\bm{K}^{\bm{x}})^{-1})\vvec(\bm{Y}),\right.\\
    &\phantom{= \mathcal{N}aa}\left. (k^{\tasks}\otimes \partial^2\bm{K}^{\bm{x}}) - (k^{\tasks\top}\otimes\partial \bm{K}^{\bm{x}\top}) ((k^{\tasks})^{-1}\otimes (\bm{K}^{\bm{x}})^{-1})(k^{\tasks}\otimes\partial \bm{K}^{\bm{x}})\right)\nonumber\\
    &= \mathcal{N}\left((\bm{I}_M\otimes\partial \bm{K}^{\bm{x}\top}(\bm{K}^{\bm{x}})^{-1})\vvec(\bm{Y}), k^{\tasks} \otimes (\partial^2 \bm{K}^{\bm{x}} - \partial \bm{K}^{\bm{x}\top} (\bm{K}^{\bm{x}})^{-1}\partial \bm{K}^{\bm{x}})\right) \\
    &= \mathcal{N}\left(\vvec(\partial \bm{K}^{\bm{x}\top}(\bm{K}^{\bm{x}})^{-1}\bm{Y}), k^{\tasks} \otimes (\partial^2 \bm{K}^{\bm{x}} - \partial \bm{K}^{\bm{x}\top} (\bm{K}^{\bm{x}})^{-1}\partial \bm{K}^{\bm{x}})\right), \label{eq:appendix:jacobian_distribution_vectorized}
\end{align}
with~\eqref{eq:appendix:jacobian_distribution_vectorized} corresponding to the posterior~\eqref{eq:method:vec_jacobian_distribution} of the main text.
Note that we used properties of matrix vectorization, and the fact that the Kronecker product behaves well with respect to matrix multiplication and inversion. Equivalently, we can formulate this distribution as a matrix-valued normal~\citep[Chap. 2.]{Gupta_Nagar:matrix_valued_dist:1999} as in Eqs.~\eqref{eq:method:jacobian_mean_posterior}-\eqref{eq:method:jacobian_cov_over_columns}.

\section{Explicit computations for multitask kernels' Jacobians}
\label{sec:appendix:matrix_computations_of_multitask_kernel_derivatives}
In this section, we explicitly compute the expressions~\eqref{eq:appendix:gram_matrix}-\eqref{eq:appendix:partial_partial_deriv_matrix}.
To do so, we consider the joint distribution of the vectors $\vvec(\bm{Y})$ and $\vvec(\bm{J}_{\feuc(\bm{x}_*)}^\top)$ in Eq.~\eqref{eq:appendix:joint_distribution_of_data_and_jacobian}. As previously mentioned, the block $\bm{K}$ is given by the covariances among the data, where we denoted $y_{mn} = y_m(\bm{x}_n)$,
\begin{equation*}
    \bm{K} = \begin{bmatrix}
    \cov(y_{11}, y_{11}) & \cdots & \cov(y_{11}, y_{N1}) & \bm{\cdots} & \cov(y_{11}, y_{1M}) & \cdots &  \cov(y_{11}, y_{NM})\\[0.1cm]
    \vdots & \ddots & \vdots & \bm{\cdots} & \vdots & \ddots & \vdots \\[0.1cm]
    \cov(y_{N1}, y_{11}) & \cdots & \cov(y_{N1}, y_{N1}) & \bm{\cdots} & \cov(y_{N1}, y_{1M}) & \cdots &  \cov(y_{N1}, y_{NM})\\[0.15cm]
    \bm{\vdots} & \bm{\vdots} & \bm{\vdots} & \bm{\cdots} & \bm{\vdots} & \bm{\vdots} & \bm{\vdots} \\[0.15cm]
    \cov(y_{1M}, y_{11}) & \cdots & \cov(y_{1M}, y_{N1}) & \bm{\cdots} & \cov(y_{1M}, y_{1M}) & \cdots & \cov(y_{1M}, y_{NM}) \\[0.1cm]
    \vdots & \ddots & \vdots & \bm{\cdots} & \vdots & \ddots & \vdots \\[0.1cm]
    \cov(y_{NM}, y_{11}) & \cdots & \cov(y_{NM}, y_{N1}) & \bm{\cdots} & \cov(y_{NM}, y_{1M}) & \cdots & \cov(y_{NM}, y_{NM})
    \end{bmatrix}.
\end{equation*}
For our multitask GP with multitask kernel $k=k^{\tasks}\otimes k^{\bm{x}}$, this covariance is explicitly given by,
\begin{equation*}
    \bm{K} = \begin{bmatrix}
    k^{\tasks}_{11}k^{\bm{x}}(\bm{x}_1, \bm{x}_1) & \cdots & k^{\tasks}_{11}k^{\bm{x}}(\bm{x}_1, \bm{x}_N) & \bm{\cdots} & k^{\tasks}_{1M}k^{\bm{x}}(\bm{x}_1, \bm{x}_1) & \cdots &  k^{\tasks}_{1M}k^{\bm{x}}(\bm{x}_1, \bm{x}_N)\\[0.1cm]
    \vdots & \ddots & \vdots & \bm{\cdots} & \vdots & \ddots & \vdots \\[0.1cm]
    k^{\tasks}_{11}k^{\bm{x}}(\bm{x}_N, \bm{x}_1) & \cdots & k^{\tasks}_{11}k^{\bm{x}}(\bm{x}_N, \bm{x}_N) & \bm{\cdots} & k^{\tasks}_{1M}k^{\bm{x}}(\bm{x}_N, \bm{x}_1) & \cdots &  k^{\tasks}_{1M}k^{\bm{x}}(\bm{x}_N, \bm{x}_N)\\[0.15cm]
    \bm{\vdots} & \bm{\vdots} & \bm{\vdots} & \bm{\cdots} & \bm{\vdots} & \bm{\vdots} & \bm{\vdots} \\[0.15cm]
    k^{\tasks}_{M1}k^{\bm{x}}(\bm{x}_1, \bm{x}_1) & \cdots & k^{\tasks}_{M1}k^{\bm{x}}(\bm{x}_1, \bm{x}_N) & \bm{\cdots} & k^{\tasks}_{MM}k^{\bm{x}}(\bm{x}_1, \bm{x}_1) & \cdots &  k^{\tasks}_{MM}k^{\bm{x}}(\bm{x}_1, \bm{x}_N)\\[0.1cm]
    \vdots & \ddots & \vdots & \bm{\cdots} & \vdots & \ddots & \vdots \\[0.1cm]
    k^{\tasks}_{M1}k^{\bm{x}}(\bm{x}_N, \bm{x}_1) & \cdots & k^{\tasks}_{M1}k^{\bm{x}}(\bm{x}_N, \bm{x}_N) & \bm{\cdots} & k^{\tasks}_{MM}k^{\bm{x}}(\bm{x}_N, \bm{x}_1) & \cdots &  k^{\tasks}_{MM}k^{\bm{x}}(\bm{x}_N, \bm{x}_N)
\end{bmatrix}.
\end{equation*}
Notice that this matrix is composed of blocks $k^{\tasks}_{ij}[k^{\bm{x}}(\bm{x}_a, \bm{x}_b)]_{a,b=1}^N$. In other words, $\bm{K}$ can equivalently be written as $\bm{K} = k^{\tasks}\otimes \bm{K}^{\bm{x}}$.

Next, we focus on $\partial \bm{K}$. Denoting the partial derivatives $\partial y_m(\bm{x}_*) / \partial x^{(r)}$ as $\partial_r y_{m*}$, we obtain,
\begin{align*}
    \label{eq:appendix:partial_K}
    \partial \bm{K} &= \begin{bmatrix}
        \cov(y_{11}, \partial_1 y_{1*}) & \cdots & \cov(y_{11}, \partial_d y_{1*}) & \bm{\cdots} & \cov(y_{11}, \partial_1 y_{M*}) & \cdots & \cov(y_{11}, \partial_d y_{M*}) \\[0.1cm]
        \vdots & \ddots & \vdots & \bm{\cdots} & \vdots & \ddots & \vdots \\[0.1cm]
        \cov(y_{N1}, \partial_1 y_{1*}) & \cdots & \cov(y_{N1}, \partial_d y_{1*}) & \bm{\cdots} & \cov(y_{N1}, \partial_1 y_{M*}) & \cdots & \cov(y_{N1}, \partial_d y_{M*}) \\[0.15cm]
        \bm{\vdots} & \bm{\vdots} & \bm{\vdots} & \bm{\cdots} & \bm{\vdots} & \bm{\vdots} & \bm{\vdots} \\[0.15cm]     
        \cov(y_{1M}, \partial_1 y_{1*}) & \cdots & \cov(y_{1M}, \partial_d y_{1*}) & \bm{\cdots} & \cov(y_{1M}, \partial_1 y_{M*}) & \cdots & \cov(y_{1M}, \partial_d y_{M*}) \\[0.1cm]
        \vdots & \ddots & \vdots & \bm{\cdots} & \vdots & \ddots & \vdots \\[0.1cm]
        \cov(y_{NM}, \partial_1 y_{1*}) & \cdots & \cov(y_{NM}, \partial_d y_{1*}) & \bm{\cdots} & \cov(y_{NM}, \partial_1 y_{M*}) & \cdots & \cov(y_{NM}, \partial_d y_{M*})
    \end{bmatrix} \\[0.3cm]
    &= \begin{bmatrix}
        \pderiv{k_{11}(\bm{x}_1, \bm{x}_*)}{x^{(1)}} & \cdots & \pderiv{k_{11}(\bm{x}_1, \bm{x}_*)}{x^{(d)}} & \bm{\cdots} & \pderiv{k_{1M}(\bm{x}_1, \bm{x}_*)}{x^{(1)}} & \cdots & \pderiv{k_{1M}(\bm{x}_1, \bm{x}_*)}{x^{(d)}} \\[0.1cm]
        \vdots & \ddots & \vdots & \bm{\cdots} & \vdots & \ddots & \vdots \\[0.1cm]
        \pderiv{k_{11}(\bm{x}_N, \bm{x}_*)}{x^{(1)}} & \cdots & \pderiv{k_{11}(\bm{x}_N, \bm{x}_*)}{x^{(d)}} & \bm{\cdots} & \pderiv{k_{1M}(\bm{x}_N, \bm{x}_*)}{x^{(1)}} & \cdots & \pderiv{k_{1M}(\bm{x}_N, \bm{x}_*)}{x^{(d)}} \\[0.15cm]
        \bm{\vdots} & \bm{\vdots} & \bm{\vdots} & \bm{\cdots} & \bm{\vdots} & \bm{\vdots} & \bm{\vdots} \\[0.15cm]
        \pderiv{k_{M1}(\bm{x}_1, \bm{x}_*)}{x^{(1)}} & \cdots & \pderiv{k_{M1}(\bm{x}_1, \bm{x}_*)}{x^{(d)}} & \bm{\cdots} & \pderiv{k_{MM}(\bm{x}_1, \bm{x}_*)}{x^{(1)}} & \cdots & \pderiv{k_{MM}(\bm{x}_1, \bm{x}_*)}{x^{(d)}} \\[0.1cm]
        \vdots & \ddots & \vdots & \bm{\cdots} & \vdots & \ddots & \vdots \\[0.1cm]
        \pderiv{k_{M1}(\bm{x}_N, \bm{x}_*)}{x^{(1)}} & \cdots & \pderiv{k_{M1}(\bm{x}_N, \bm{x}_*)}{x^{(d)}} & \bm{\cdots} & \pderiv{k_{MM}(\bm{x}_N, \bm{x}_*)}{x^{(1)}} & \cdots & \pderiv{k_{MM}(\bm{x}_N, \bm{x}_*)}{x^{(d)}}
    \end{bmatrix} \\[0.3cm]
    &= \begin{bmatrix}
        k_{11}^{\tasks}\pderiv{k^{\bm{x}}(\bm{x}_1, \bm{x}_*)}{x^{(1)}} & \cdots & k_{11}^{\tasks} \pderiv{k^{\bm{x}}(\bm{x}_1, \bm{x}_*)}{x^{(d)}} & \bm{\cdots} & k_{1M}^{\tasks}\pderiv{k^{\bm{x}}(\bm{x}_1, \bm{x}_*)}{x^{(1)}} & \cdots & k_{1M}^{\tasks} \pderiv{k^{\bm{x}}(\bm{x}_1, \bm{x}_*)}{x^{(d)}} \\[0.1cm]
        \vdots & \ddots & \vdots & \bm{\cdots} & \vdots & \ddots & \vdots \\[0.1cm]
        k_{11}^{\tasks}\pderiv{k^{\bm{x}}(\bm{x}_N, \bm{x}_*)}{x^{(1)}} & \cdots & k_{11}^{\tasks} \pderiv{k^{\bm{x}}(\bm{x}_N, \bm{x}_*)}{x^{(d)}} & \bm{\cdots} & k_{1M}^{\tasks}\pderiv{k^{\bm{x}}(\bm{x}_N, \bm{x}_*)}{x^{(1)}} & \cdots & k_{1M}^{\tasks} \pderiv{k^{\bm{x}}(\bm{x}_N, \bm{x}_*)}{x^{(d)}} \\[0.15cm]
        \bm{\vdots} & \bm{\vdots} & \bm{\vdots} & \bm{\cdots} & \bm{\vdots} & \bm{\vdots} & \bm{\vdots} \\[0.15cm]
        k_{M1}^{\tasks}\pderiv{k^{\bm{x}}(\bm{x}_1, \bm{x}_*)}{x^{(1)}} & \cdots & k_{M1}^{\tasks} \pderiv{k^{\bm{x}}(\bm{x}_1, \bm{x}_*)}{x^{(d)}} & \bm{\cdots} & k_{MM}^{\tasks}\pderiv{k^{\bm{x}}(\bm{x}_1, \bm{x}_*)}{x^{(1)}} & \cdots & k_{MM}^{\tasks} \pderiv{k^{\bm{x}}(\bm{x}_1, \bm{x}_*)}{x^{(d)}} \\[0.1cm]
        \vdots & \ddots & \vdots & \bm{\cdots} & \vdots & \ddots & \vdots \\[0.1cm]
        k_{M1}^{\tasks}\pderiv{k^{\bm{x}}(\bm{x}_N, \bm{x}_*)}{x^{(1)}} & \cdots & k_{M1}^{\tasks} \pderiv{k^{\bm{x}}(\bm{x}_N, \bm{x}_*)}{x^{(d)}} & \bm{\cdots} & k_{MM}^{\tasks}\pderiv{k^{\bm{x}}(\bm{x}_N, \bm{x}_*)}{x^{(1)}} & \cdots & k_{MM}^{\tasks} \pderiv{k^{\bm{x}}(\bm{x}_N, \bm{x}_*)}{x^{(d)}}
    \end{bmatrix}.
\end{align*}
As previously shown, we can equivalently write $\partial \bm{K}= k^{\tasks}\otimes \partial \bm{K}^{\bm{x}}$, with $\partial \bm{K}^{\bm{x}} = [\partial k^{\bm{x}}(\bm{x}_n, \bm{x}_*) / \partial x^{(r)}]_{n,r=1}^{N,d}$.
The computations for $\partial^2 \bm{K}$ are analogous, and are left to the reader.

\section{Tangent space representations}
\label{app:tangentspaces}
This section elaborates on the choice of tangent space representations for the manifolds considered in our experiments. As discussed in Section~\ref{sec:method:vectorization_and_practicalities}, instead of considering tangent vectors in the ambient space in which $\manifold$ is embedded, we specify them with respect to a basis of the tangent space itself. This process can be viewed as a change of basis, as explained next.

\subsection{Sphere}
The sphere $\sphere^M$ is embedded in the Euclidean space $\euclideanspace^{M+1}$. Each point $\bm{p}\in\sphere^M$ is described, in coordinates, as a ($M+1$)-dimensional unit-norm vector. Each tangent vector $\bm{v}\in\mathcal{T}_{\bm{p}}\sphere^M$ is expressed as a ($M+1$)-dimensional vector orthogonal to $\bm{p}$, i.e., $\langle \bm{p}, \frac{\bm{v}}{||\bm{v}||}\rangle = 0$. Principal operations on the sphere are given by Table~\ref{tab:ManifoldOperations}-\emph{left}. When considering WGPs, we express tangent vectors in a $M$-dimensional local basis of the tangent space $\mathcal{T}_{\bm{x}}\sphere^M$. To do so, we define a basis $\bm{B}_{\bm{p}}=[\bm{b}_{\bm{p}}^1 \ldots \bm{b}_{\bm{p}}^M]$ composed of $M$ orthonormal tangent vectors $\bm{b}_{\bm{p}}^i \in\euclideanspace^{M+1}$. Tangent vectors $\bm{v}_{\bm{p}}$ in the local basis and their counterparts $\bm{v}$ in the ambient basis are consequently related as,
\begin{equation}
    \bm{v}_{\bm{p}} = \bm{B}_{\bm{p}}^\top \bm{v} \quad\quad \text{and} \quad\quad \bm{v} = \bm{B}_{\bm{p}} \bm{v}_{\bm{p}}.
\end{equation}
In the WGPs, we use the former change of basis to define the output of $\Log_{b(\bm{x})}(\bm{q})$ locally in the tangent space $\mathcal{T}_{b(\bm{x})}\sphere^M$. The latter is then used to express the posterior mean of the Euclidean GP, computed in the local basis, in the ambient basis before projecting it onto the manifold with the exponential map $\Exp_{b(\bm{x})}(\bm{q})$. Next, we discuss the choice of basis $\bm{B}_{\bm{p}}$ for $\sphere^2$ and $\sphere^3$.

\textbf{Tangent basis for $\sphere^2$:}
Denoting each point $\bm{p}\in\sphere^2$ in coordinates as $\bm{p}=(x \: y \: z)^\top$, we define the orthonormal basis $\bm{B}_{\bm{x}}$ via the reduced QR decomposition,
\begin{equation}
    \bm{A}_{\bm{x}} = \bm{B}_{\bm{x}} \bm{R}_{\bm{x}}  \quad\quad \text{with} \quad\quad \bm{A}_{\bm{x}} = \left( \begin{matrix}
        -z & 0 \\
        0 & z \\
        x & y 
    \end{matrix} \right).
\end{equation}

\textbf{Tangent basis for $\sphere^3$:} Denoting each point $\bm{p}\in\sphere^3$ in coordinates as $\bm{p}=(w \: x \: y \: z)^\top$, we define the orthonormal basis for each tangent space $\mathcal{T}_{b(\bm{x})}\sphere^3$ as,
\begin{equation}
    \bm{B}_{\bm{x}} = \left( \begin{matrix}
        -x & -y & -z \\
        w & z & -y \\
        -z & w & x \\
        y & -x & w 
    \end{matrix} \right).
\end{equation}

\subsection{SPD manifold}
The SPD manifold $\SPD^M$ is embedded in the space of symmetric matrices $\text{Sym}^M$, which is itself embedded in the Euclidean space. Each point $\bm{P}\in\SPD^M$ is a $M\times M$ symmetric positive-define matrix and each tangent vector $\bm{V}\in\mathcal{T}_{\bm{P}}\SPD^M$ is expressed as a $M\times M$ symmetric matrix. Principal operations on the SPD manifold are given by Table~\ref{tab:ManifoldOperations}-\emph{right}. When considering WGPs, we express tangent vectors in a $M(M+1)/2$-dimensional local basis of the tangent space $\mathcal{T}_{\bm{P}}\SPD^M$. This is simply achieved by defining tangent vectors $\bm{v}_{\bm{P}}$ in the local basis as vectors composed of the diagonal and upper triangular elements of $\bm{V}$.

\begin{table}
	\begin{center}
     \begin{tabular}{l|c}
		\textbf{Operation} & \textbf{Formula on $\sphere^M$} \\ [0.3ex]
		\hline 
		$\langle \bm{u}, \bm{v} \rangle_{\bm{x}}$ & $\langle \bm{u}, \bm{v} \rangle$ \\ [0.3ex] 
		\hline 
		$d_{\sphere^M}(\bm{x},\bm{y})$ & $\arccos(\bm{x}^\trsp \bm{y})$ \\ [0.3ex] 
		\hline
		$\Exp_{\bm{x}}(\bm{u})$ & $\bm{x}\cos(\|\bm{u}\|) + \frac{\bm{u}}{\|\bm{u}\|}\sin(\|\bm{u}\|)$\\ [0.3ex]
		\hline 
		$\Log_{\bm{x}}(\bm{u})$ & $ d_{\sphere^M}(\bm{x},\bm{y}) \, \frac{\bm{y} - \bm{x}^\trsp \bm{y} \, \bm{x}}{\|\bm{y} - \bm{x}^\trsp \bm{y} \, \bm{x}\|}$\\ [0.3ex]
	\end{tabular}
 \hspace{1cm}
	\begin{tabular}{l|c}
		\textbf{Operation} & \textbf{Formula on $\SPD^M$} \\ [0.3ex]
		\hline 
		$\langle \bm{U}, \bm{V} \rangle_{\bm{X}}$ & $\mathrm{tr}\left(\bm{X}^{-\frac{1}{2}}\bm{U}\bm{X}^{-1}\bm{V}\bm{X}^{-\frac{1}{2}}\right)$ \\ [0.3ex] 
		\hline 
		$d_{\SPD^M}(\bm{X},\bm{Y})$ & $\|\log(\bm{X}^{-\frac{1}{2}}\bm{Y}\bm{X}^{-\frac{1}{2}})\|_\text{F}$ \\ [0.3ex] 
		\hline
		$\Exp_{\bm{X}}(\bm{U})$ & $\bm{X}^{\frac{1}{2}}\exp(\bm{X}^{-\frac{1}{2}}\bm{U}\bm{X}^{-\frac{1}{2}})\bm{X}^{\frac{1}{2}}$\\ [0.3ex]
		\hline 
		$\Log_{\bm{X}}(\bm{Y})$ & $\bm{X}^{\frac{1}{2}}\log(\bm{X}^{-\frac{1}{2}}\bm{Y}\bm{X}^{-\frac{1}{2}})\bm{X}^{\frac{1}{2}}$\\ [0.3ex]
	\end{tabular}
 \end{center}
 \caption{\emph{Left:} Principal operations on $\sphere{d}$, see~\citep{Absil07} or~\citep{boumal2020intromanifolds} for details. \emph{Right:} Principal operations on $\SPD^M$ when endowed with the affine-invariant metric, see~\citep{Pennec06:AIM} for details.}
    \label{tab:ManifoldOperations}
\end{table}

\section{Likelihood of Wrapped Gaussian Processes}
\label{app:wgp}
A WGP is $\Exp_{b(\cdot)}(\feuc(\cdot))$, where $\feuc\sim\GP(\bm{0}, k)$ is a Euclidean GP learned on the tangent spaces $\mathcal{T}_{b(\cdot)}\manifold$. In other words, the exponential map pushes forward the Euclidean distribution onto the manifold. Therefore, the WGP marginal likelihood must account for the change of volume induced by the exponential map. This is achieved by leveraging the change of variable formula.
Specifically, the change of variable formula states that, given a random variable $\bm{V}$ endowed with the probability function $p(\bm{v})$, the log-likelihood of $\bm{Y}=g(\bm{V})$ at $\bm{y}$ is expressed as,
\begin{equation}
\label{eq:appendix:change_of_variable}
    \log p(\bm{y}) = \log p(\bm{v}) - \log\det \frac{\partial g}{\partial \bm{v}}.
\end{equation}
In the case of a WGP, $p(\bm{x})$ is equal to the probability function of the Euclidean GP $\feuc$ and the function $g$ is the exponential map $\Exp_{b(\cdot)}$. Therefore, as discussed in Section~\ref{sec:method:training}, the WGP marginal likelihood is,
\begin{equation*}
    \log p(\bm{y}|\bm{x}) = \log \mathcal{N}\big(\bm{v}; \bm{0}, \bm{K}\big) - \log \det \left(\frac{\partial \Exp_{b(\bm{x})}}{\partial \bm{v}}\right),
\end{equation*}
with tangent vectors $\bm{v}$. Next, we provide the expression of the term $\log \det \left(\frac{\partial \Exp_{b(\bm{x})}}{\partial \bm{v}}\right)$ induced by the change of variable for the manifolds used in our experiments.

\subsection{Sphere}
The exponential map for the sphere manifold $\sphere^M$ is given by,
\begin{equation}
    \Exp_{\bm{p}}(\bm{v}) = \bm{p}\cos(\|\bm{v}\|) + \frac{\bm{v}}{\|\bm{v}\|}\sin(\|\bm{v}\|),
\end{equation}
leading to the change of variable correction
\begin{equation}
    \log \det \left(\frac{\partial \Exp_{\bm{x}}(\bm{v})}{\partial \bm{v}}\right) = (M-1) \log \left( \frac{\sin^2(||\bm{v}||)}{||\bm{v}||^2}\right).
\end{equation}

\subsection{SPD manifold}
The exponential map for the SPD manifold $\SPD^M$ endowed with the affine-invariant Riemannian metric~\citep{Pennec06:AIM} is given by,
\begin{equation}
    \Exp_{\bm{P}}(\bm{V}) = \bm{X}^{\frac{1}{2}}\exp(\bm{X}^{-\frac{1}{2}}\bm{V}\bm{X}^{-\frac{1}{2}})\bm{X}^{\frac{1}{2}},
\end{equation}
where $\exp$ denotes the matrix exponential.
In our experiments, we consider the basepoints of the WGP to be multiples of the identity matrix, i.e., $\bm{P} = b(\bm{X}) = a\bm{I}$. In this case, the exponential map simplifies as,
\begin{equation}
    \Exp_{a\bm{I}}(\bm{V}) = a \exp(a^{-1}\bm{V}),
\end{equation}
leading to the change of variable correction
\begin{equation}
    \log \det \left(\frac{\partial \Exp_{a\bm{I}}(\bm{V})}{\partial \bm{V}}\right) = \exp(a^{-1}\bm{V}).
\end{equation}

\section{Experimental Details}
\label{app:experimental_details}

We implemented all approaches using \texttt{GPyTorch}~\citep{gardner2018gpytorch}. 
We used the \texttt{Geomstats} implementation of the exponential and logarithmic maps~\citep{miolane:geomstats:2020}, with minor modifications for the vectorization process discussed in Sec.~\ref{sec:method:vectorization_and_practicalities}. 
To compute geodesics in latent spaces we used \texttt{StochMan}~\citep{software:stochman}, and the robotic simulations rely on the \texttt{Python Robotics Toolbox}~\citep{corke:robotictoolbox:2021}.
When back-constrained models are necessary, the Riemannian kernels are implemented using the \texttt{GeometricKernels} Python package~\citep{Mostowsky24:GeomKernels}.

We used SE kernels $k^{\bm{x}}$ for all models and constant basepoint functions $b(\bm{x})=\bm{p}$ for WGPLVM and \riemannsq, with $\bm{p}=(1, 0, \ldots, 0)^\trsp$ for $\sphere^M$ and $\bm{p}=\bm{I}$ for $\SPD^M$. The tangent space vectors $\bm{v}_i=\Log_{b(\bm{x}_i)}(\bm{y}_i)$ were centered in order to build a zero-mean Euclidean GP $\feuc$. The latent variables were initialized using PCA. The latent variables and GP parameters were optimized by minimizing the marginal likelihood for $1000$ iterations using Adam~\citep{KingmaBa15:Adam}. 
Data and experiment-specific parameters are detailed next. 

\subsection{Illustrative example on $\euclideanspace^2 \times \sphere^2$}
The dataset of this experiment is composed of $6$ trajectories of $200$ datapoints. Each trajectory traces a $\mathsf{J}$ shape in $\euclideanspace^2$ and a $\mathsf{C}$ shape on the $\sphere^2$.
All models are augmented with a GPDM latent prior with dynamic kernel defined as a SE kernel with learnable lengthscale and variance. The learning rate of all models is fixed at $0.025$.
The geodesics are computed by discretizing the 2D latent manifold into a $50\times 50$ graph and computing the shortest path on the obtained graph via classical algorithms using \texttt{StochMan}~\citep{software:stochman}.

\subsection{Robot motion synthesis on $\euclideanspace^3 \times \sphere^3$}
The dataset of this experiment is composed of $6$ trajectories of varying length ($60$ to $85$ datapoints) for a total of $607$ datapoints. We use a Gamma prior with concentration $\alpha=2$ and rate $\beta=2$ for the GP kernel lengthscale of all models. We augment the models with back constraints. For GPLVM and pGPLVM, we use a Euclidean SE back-constraints kernel $k^{\manifold}=k^{\euclideanspace^7}$ with lengthscale $\theta=0.2$ and variance $\sigma^2=1$. For WGPLVM and \riemannsq, we use a back-constraints kernel defined as the product $k^{\manifold}=k^{\euclideanspace^3} k^{\sphere^3}$. The Euclidean kernel $k^{\euclideanspace^3}$ is a SE kernel with lengthscale $\theta=0.2$ and variance $\sigma^2=1$. The sphere kernel $k^{\sphere^3}$ is a sphere SE kernel formulated as in~\citep{Borovitskiy20:GPManifolds} with lengthscale $\theta=0.2$ and variance $\sigma^2=1$.
The learning rate of all models is fixed at $0.05$.
For all models, geodesics are parametrized by cubic splines, whose parameters are optimized to minimize the curve energy using \texttt{StochMan}~\citep{software:stochman}. Specifically, we approximate the geodesics by cubic splines $c \approx {\omega_\lambda}(\bm{z}_{c})$,
with $\bm{z}_c = \{\bm{z}_{c_0}, \ldots, \bm{z}_{c_N} \}$, where $\bm{z}_{c_n} \in \mathbb{R}^Q$ is a vector defining a control point of the spline over the latent space. Given $N$ control points, $N-1$ cubic polynomials $\omega_{\lambda_i}$ with coefficients $\lambda_{i,0}$, $\lambda_{i,1}$, $\lambda_{i,2}$, $\lambda_{i,3}$ have to be estimated to minimize the curve energy.

\subsection{Manipulability learning in $\euclideanspace^2 \times \SPD^2$}
The dataset of this experiment is composed of $6$ trajectories of $100$ datapoints. Each trajectory traces a $\mathsf{C}$ shape in $\euclideanspace^2$. The manipulability profiles correspond to the manipulability of a planar robot whose end-effector follows this $\mathsf{C}$ shape.
We use a Gamma prior with concentration $\alpha=2$ and rate $\beta=2$ for the WGP kernel lengthscale of WGPLVM and \riemannsq.
All models are augmented with a GPDM latent prior and with back constraints. The GPDM dynamic kernel is defined as a SE kernel with learnable lengthscale and variance for all models.
For GPLVM and pGPLVM, we use a Euclidean SE back-constraints kernel $k^{\manifold}=k^{\euclideanspace^7}$ with lengthscale $\theta=0.2$ and variance $\sigma^2=1$. For WGPLVM and \riemannsq, we use a back-constraints kernel defined as the product $k^{\manifold}=k^{\euclideanspace^2} k^{\SPD^2}$. The Euclidean kernel $k^{\euclideanspace^2}$ is a SE kernel with lengthscale $\theta=0.5$ and variance $\sigma^2=1$. The SPD kernel $k^{\SPD^2}$ is a SPD SE kernel formulated as in~\citep{Azangulov23:NonCompactKernels} with lengthscale $\theta=0.5$ and variance $\sigma^2=1$.
The learning rate of all models is fixed at $0.025$.
For all models, geodesics are parametrized by cubic splines, whose parameters are optimized to minimize the curve energy using \texttt{StochMan}~\citep{software:stochman}, similar to the previous experiment.

\subsection{Brain connectomes in $\SPD^{15}$}
We consider resting-state functional brain connectome data from the ``$1200$ Subjects release'' of the Human Connectome Project~\citep{VanEssen13:HumanConnectome}. Specifically, we consider the $200$ first subjects out of the sub-dataset ``R $812$'', which includes rs-fMRI data.
We choose the parcellation of the brain with $N = 15$ regions and use the nodetime series provided in the dataset. Specifically, for each subject, we use $15$ time series corresponding to the activation of each of the $15$ brain regions over time. We build the parcellated connectome for each subject as the $15\times 15$ covariance matrix corresponding to the correlations between nodes over the time serie. Each connectome is represented as a point in $\SPD^{15}$.
The learning rate of all models is fixed at $0.03$.
For all models, geodesics are parametrized by cubic splines, whose parameters are optimized to minimize the curve energy using \texttt{StochMan}~\citep{software:stochman}.

\subsection{Dynamic time warping distance}
We use dynamic time warping distance (DTWD) as the established quantitative measure of reconstruction accuracy of a decoded geodesic w.r.t.\@ a demonstrated trajectory, assuming equal initial conditions. The DTWD is defined as,
\begin{equation*}
    \operatorname{DTWD}({\tau_x}, \tau_{{x}^\prime}) =
     \sum_{j \in l({\tau}_{x^\prime})} \min_{i \in l(\tau_{x})} \left( d(\tau_{x_i}, {\tau}_{x_j^\prime}) \right) + \sum_{i \in l(\tau_{x})} \min_{j \in l({\tau}_{x^\prime})} \left(d(\tau_{x_i}, {\tau}_{x_j^\prime}) \right),
\end{equation*}
where ${\tau_x}$ and $\tau_{{x}^\prime}$ are two trajectories (e.g.\@ the decoded geodesic and a demonstration trajectory), $d$ is a distance function (e.g.\@ Euclidean distance), and $l({\tau})$ is the length of trajectory ${\tau}$. 

\section{Additional Results}
\label{app:additional_results}
\begin{figure*}
    \centering
    \begin{subfigure}[b]{\textwidth}
    \centering
        \includegraphics[trim={0.0cm 1.5cm 0.0cm 1.5cm},clip,width=0.2\textwidth]{Figures/R2xSPD2/wrapped_bcgpdm_on_r2_spd2_LatentSpaceMetricVol.png}
        \includegraphics[trim={0.0cm 1.5cm 0.0cm 1.5cm},clip,width=0.2\textwidth]{Figures/R2xSPD2/wrapped_bcgpdm_on_r2_spd2_R2SPD2.png}
        \includegraphics[trim={0.0cm 0.0cm 10.0cm 0.0cm},clip,width=0.15\textwidth]{Figures/R2xSPD2/wrapped_bcgpdm_on_r2_spd2_SPD2.png}
        \includegraphics[trim={5.0cm 0.3cm 5.0cm 0.5cm},clip,width=0.3\textwidth]{Figures/R2xSPD2/wrapped_bcgpdm_on_r2_spd2_SP2time.png}
        \caption{WPGLVM and \riemannsq with correlated output dimensions.}
        \label{fig:experiments:spd_supplementary:corrRiemann2}
    \end{subfigure}%

    \begin{subfigure}[b]{\textwidth}
    \centering
        \includegraphics[trim={0.0cm 1.5cm 0.0cm 1.5cm},clip,width=0.2\textwidth]{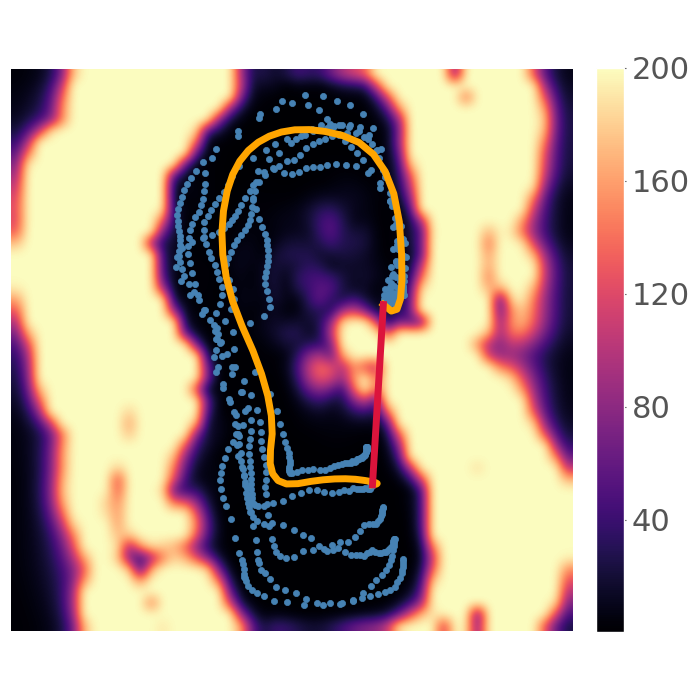}
        \includegraphics[trim={0.0cm 1.5cm 0.0cm 1.5cm},clip,width=0.2\textwidth]{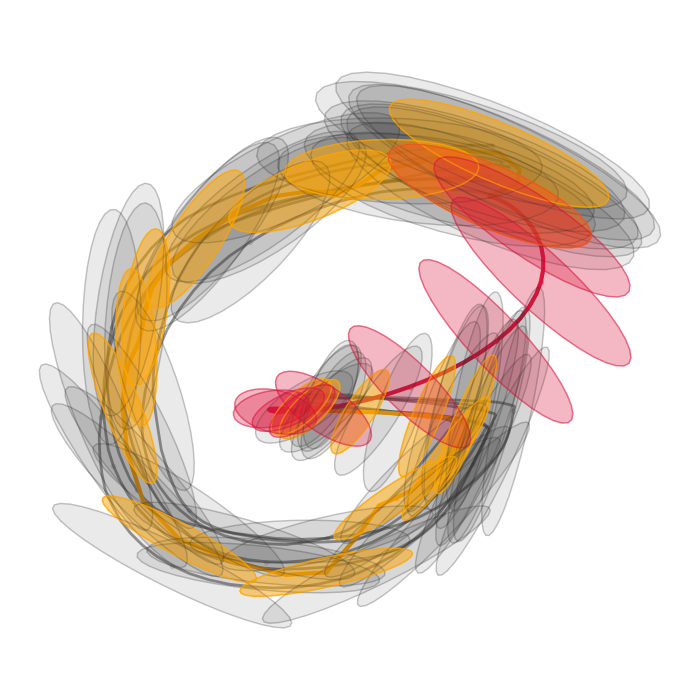}
        \includegraphics[trim={0.0cm 0.0cm 10.0cm 0.0cm},clip,width=0.15\textwidth]{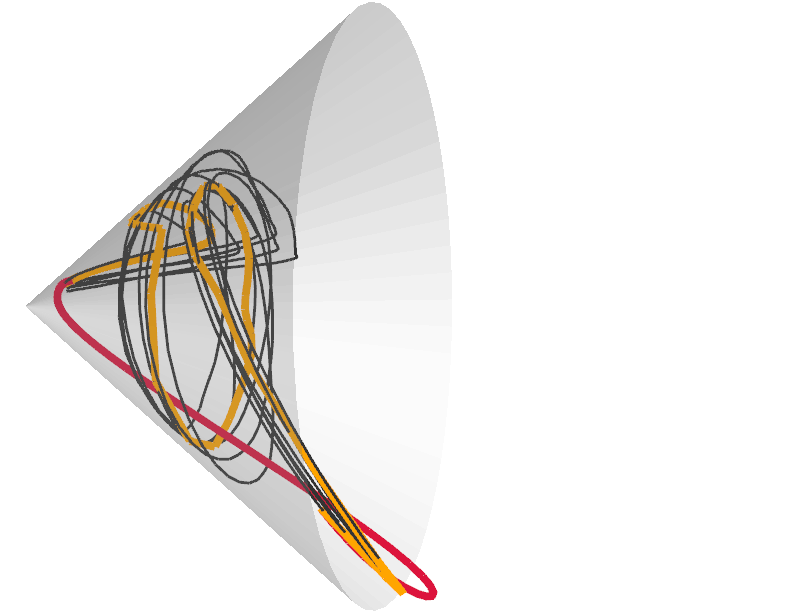}
        \includegraphics[trim={5.0cm 0.3cm 5.0cm 0.5cm},clip,width=0.3\textwidth]{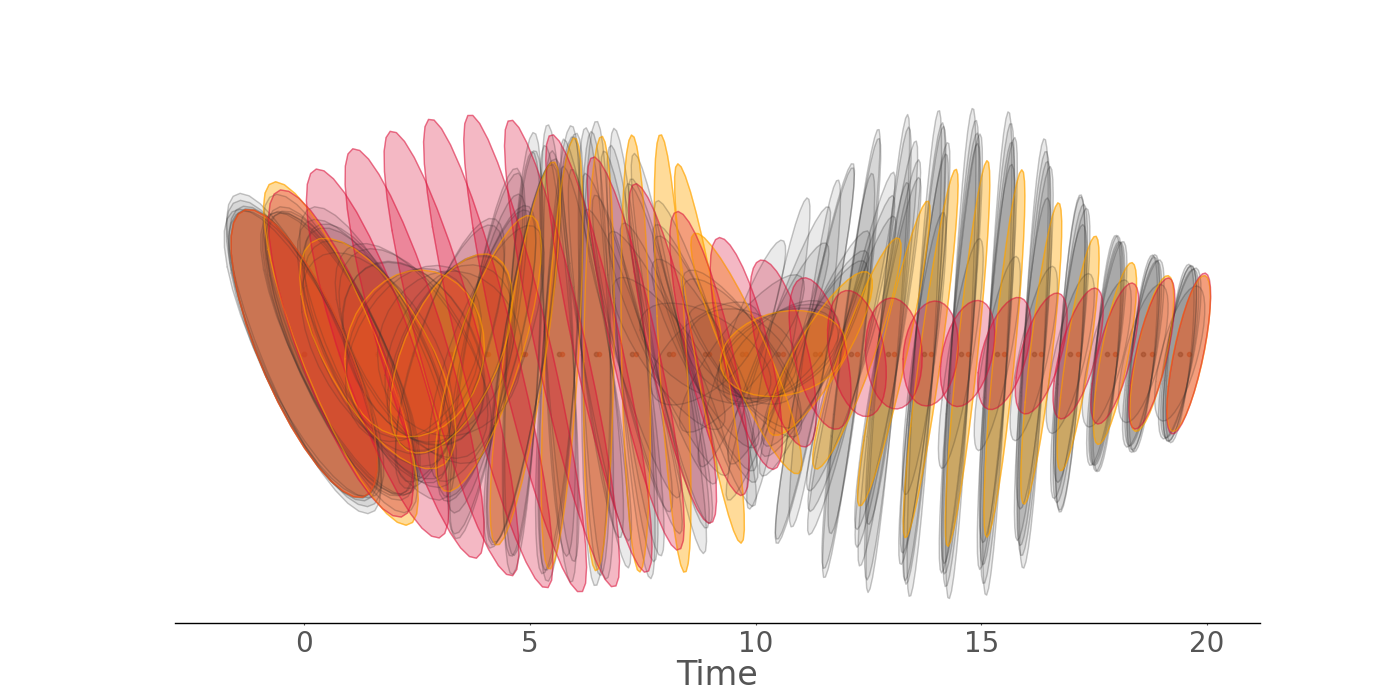}
        \caption{WPGLVM and \riemannsq with independent output dimensions.}
        \label{fig:experiments:spd_supplementary:indRiemann2}
    \end{subfigure}%
    
    \begin{subfigure}[b]{\textwidth}
    \centering
        \includegraphics[trim={0.0cm 1.5cm 0.0cm 1.5cm},clip,width=0.2\textwidth]{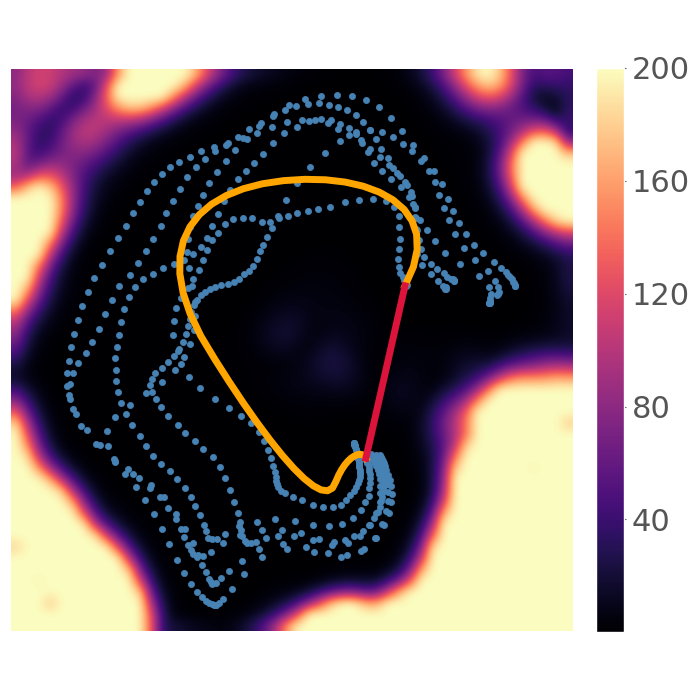}
        \includegraphics[trim={0.0cm 1.5cm 0.0cm 1.5cm},clip,width=0.2\textwidth]{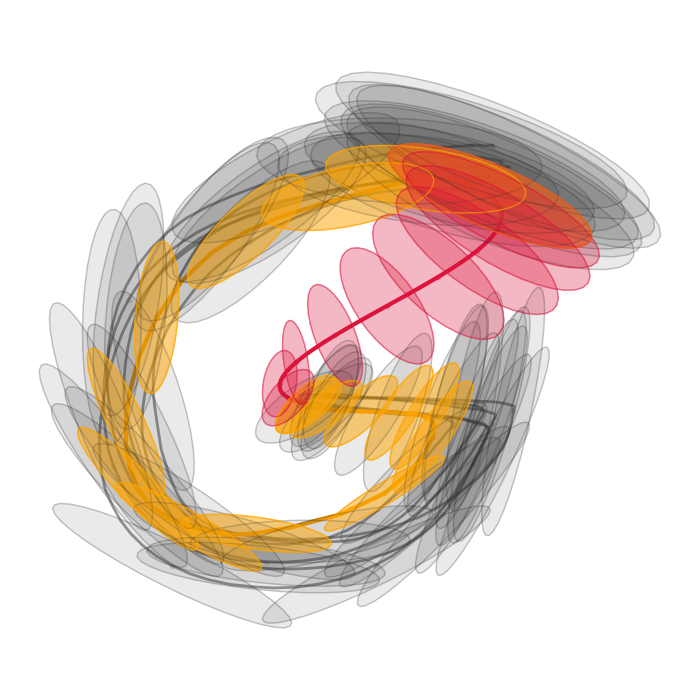}
        \includegraphics[trim={0.0cm 0.0cm 10.0cm 0.0cm},clip,width=0.15\textwidth]{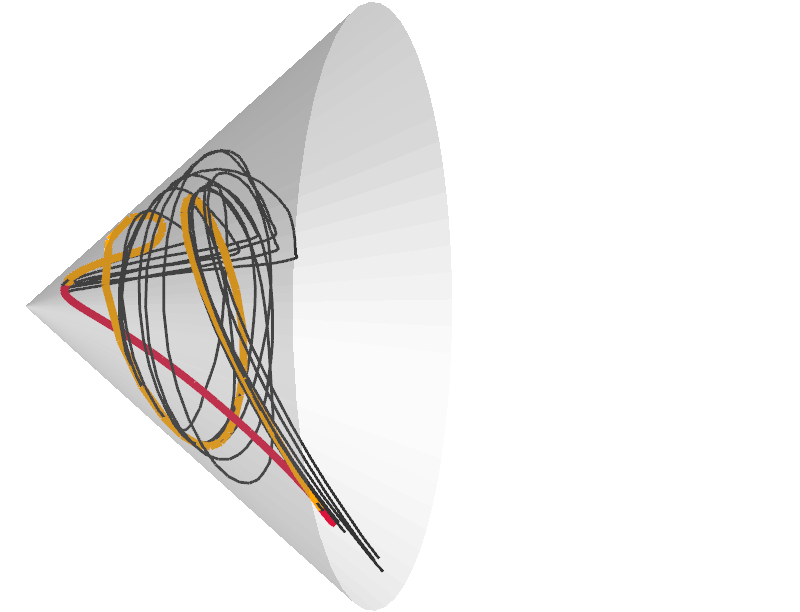}
        \includegraphics[trim={5.0cm 0.3cm 5.0cm 0.5cm},clip,width=0.3\textwidth]{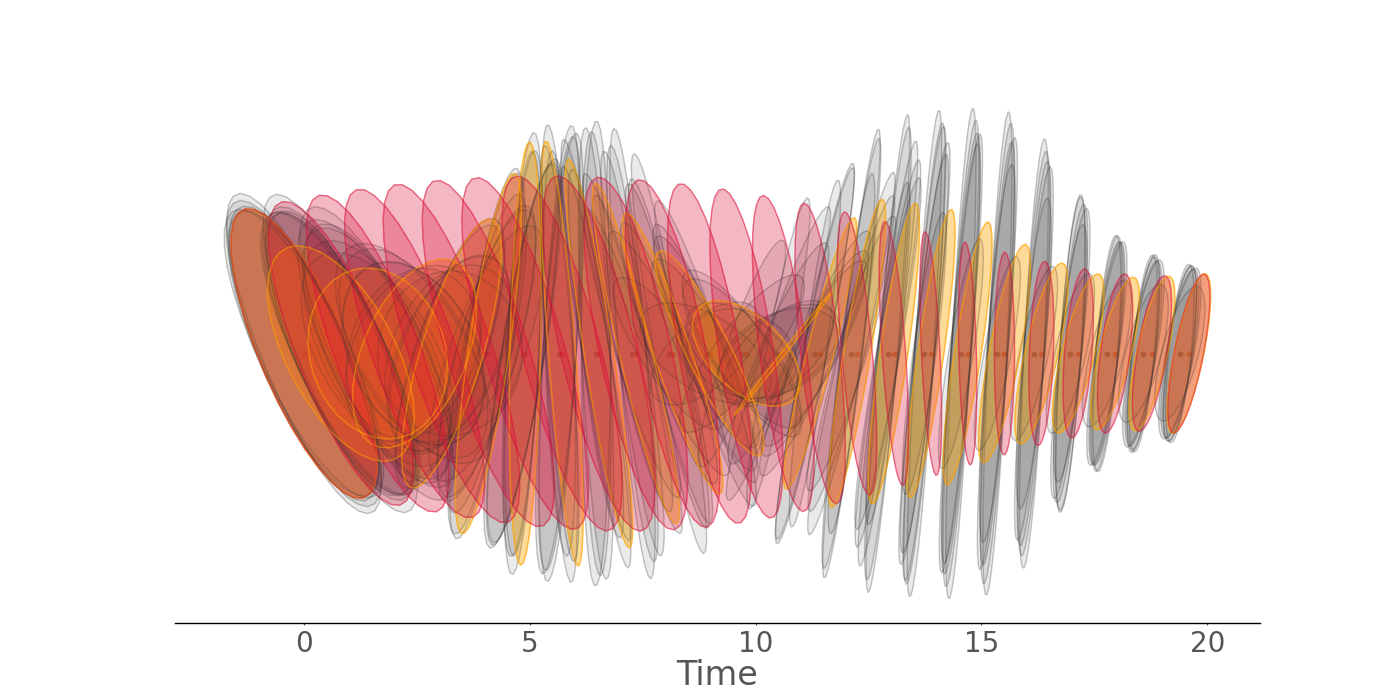}
        \caption{GPLVM and pGPLVM with correlated output dimensions.}
        \label{fig:experiments:spd_supplementary:corrGPLVM}
    \end{subfigure}%

    \begin{subfigure}[b]{\textwidth}
    \centering
        \includegraphics[trim={0.0cm 1.5cm 0.0cm 1.5cm},clip,width=0.2\textwidth]{Figures/R2xSPD2/exact_bcgpdm_on_r2_spd2_bi_LatentSpaceMetricVol.png}
        \includegraphics[trim={0.0cm 1.5cm 0.0cm 1.5cm},clip,width=0.2\textwidth]{Figures/R2xSPD2/exact_bcgpdm_on_r2_spd2_bi_R2SPD2.png}
        \includegraphics[trim={0.0cm 0.0cm 10.0cm 0.0cm},clip,width=0.15\textwidth]{Figures/R2xSPD2/exact_bcgpdm_on_r2_spd2_bi_SPD2.png}
        \includegraphics[trim={5.0cm 0.3cm 5.0cm 0.5cm},clip,width=0.3\textwidth]{Figures/R2xSPD2/exact_bcgpdm_on_r2_spd2_bi_SP2time.png}
        \caption{GPLVM and pGPLVM with independent output dimensions.}
        \label{fig:experiments:spd_supplementary:indGPLVM}
    \end{subfigure}%
    \caption{$\euclideanspace^2 \times \SPD^2$: From \emph{left} to \emph{right}: Latent variables (\dodgerbluecircle) with magnification factor of the pullback metrics, demonstrations (\blackline,\grayellipse) and reconstructions depicted as curves and ellipsoids in $\euclideanspace^2$, on the manifold $\SPD^2$, and as ellipsoids over time. One Euclidean (\crimsonline) and one Riemannian (\yellowline) geodesic are depicted for each model. }
    \label{fig:experiments:spd_supplementary}
\end{figure*}

Figure~\ref{fig:experiments:spd_supplementary} complements the results of Section~\ref{sec:experiments:manipulability} by showing the latent spaces learned with GPLVM, pGPLVM, WGPLVM, and \riemannsq with both independent and correlated output dimensions. 
As discussed in Section~\ref{sec:experiments:manipulability} and shown in Table~\ref{tab:experiments:spd_experiment}, \riemannsq with correlated output dimensions achieve the best results overall as it accounts for the correlations between positions and manipulability along the trajectories. This is particularly visible in the \emph{right-most} panel of Figures~\ref{fig:experiments:spd_supplementary:corrRiemann2} and~\ref{fig:experiments:spd_supplementary:indRiemann2}, where the decoded manipulability profile of \riemannsq with correlated output matches the demonstrations more than the manipulability profile obtained via independent output dimensions. We observe similar patterns for the pGPLVM of Figures~\ref{fig:experiments:spd_supplementary:corrGPLVM} and~\ref{fig:experiments:spd_supplementary:indGPLVM}, even if these models do not account for the geometry of SPD data.

\end{document}